\documentclass[runningheads]{llncs}

\usepackage{eccv}

\usepackage{eccvabbrv}
\usepackage{graphicx}
\usepackage{booktabs}
\usepackage{tabularx}
\usepackage{threeparttable}
\usepackage{multirow}
\usepackage{arydshln}
\usepackage{wrapfig}
\usepackage{enumitem}
\usepackage{adjustbox}
\usepackage{moresize}
\usepackage{color}

\usepackage[accsupp]{axessibility}  

\usepackage[pagebackref,breaklinks,colorlinks,citecolor=eccvblue]{hyperref}

\usepackage{orcidlink}

\begin{document}

\title{FAMOUS: High-\textcolor{blue!50!white}{F}idelity Monocul\textcolor{blue!50!white}{a}r \\ 
3D Hu\textcolor{blue!50!white}{m}an Digitizati\textcolor{blue!50!white}{o}n \textcolor{blue!50!white}{U}sing View \textcolor{blue!50!white}{S}ynthesis}

\titlerunning{3D Human Digitization}

\author{Vishnu Mani Hema\inst{1}\orcidlink{0000-0001-6489-7468} 
Shubhra Aich\inst{1}\orcidlink{0000-0002-5117-5164} 
Christian Haene\inst{2}\orcidlink{0000-0002-8599-6681} \\
Jean-Charles Bazin\inst{2}\orcidlink{0000-0001-7660-4802}
Fernando De la Torre\inst{1}\orcidlink{0000-0002-7086-8572}
}

\authorrunning{V. M. Hema, S. Aich, C. Haene, J.C. Bazin, F. Torre.}

\institute{Carnegie Mellon University \and
Independent Researcher\\
\email{vmanihem@alumni.cmu.edu}}

\maketitle

\begin{abstract}
\noindent The advancement in deep implicit modeling and articulated models has significantly enhanced the process of digitizing human figures in 3D from just a single image. While state-of-the-art methods have greatly improved geometric precision, the challenge of accurately inferring texture remains, particularly in obscured areas such as the back of a person in frontal-view images. This limitation in texture prediction largely stems from the scarcity of large-scale and diverse 3D datasets, whereas their 2D counterparts are abundant and easily accessible. To address this issue, our paper proposes leveraging extensive 2D fashion datasets to enhance both texture and shape prediction in 3D human digitization. We incorporate 2D priors from the fashion dataset to learn the occluded back view, refined with our proposed domain alignment strategy. We then fuse this information with the input image to obtain a fully textured mesh of the given person. Through extensive experimentation on standard 3D human benchmarks, we demonstrate the superior performance of our approach in terms of both texture and geometry. Code and dataset is available at \url{https://github.com/humansensinglab/FAMOUS}.
\end{abstract}

\section{Introduction}
\label{sec:intro}
\begin{figure*}[!htbp]
\centering
\includegraphics[width=\textwidth]{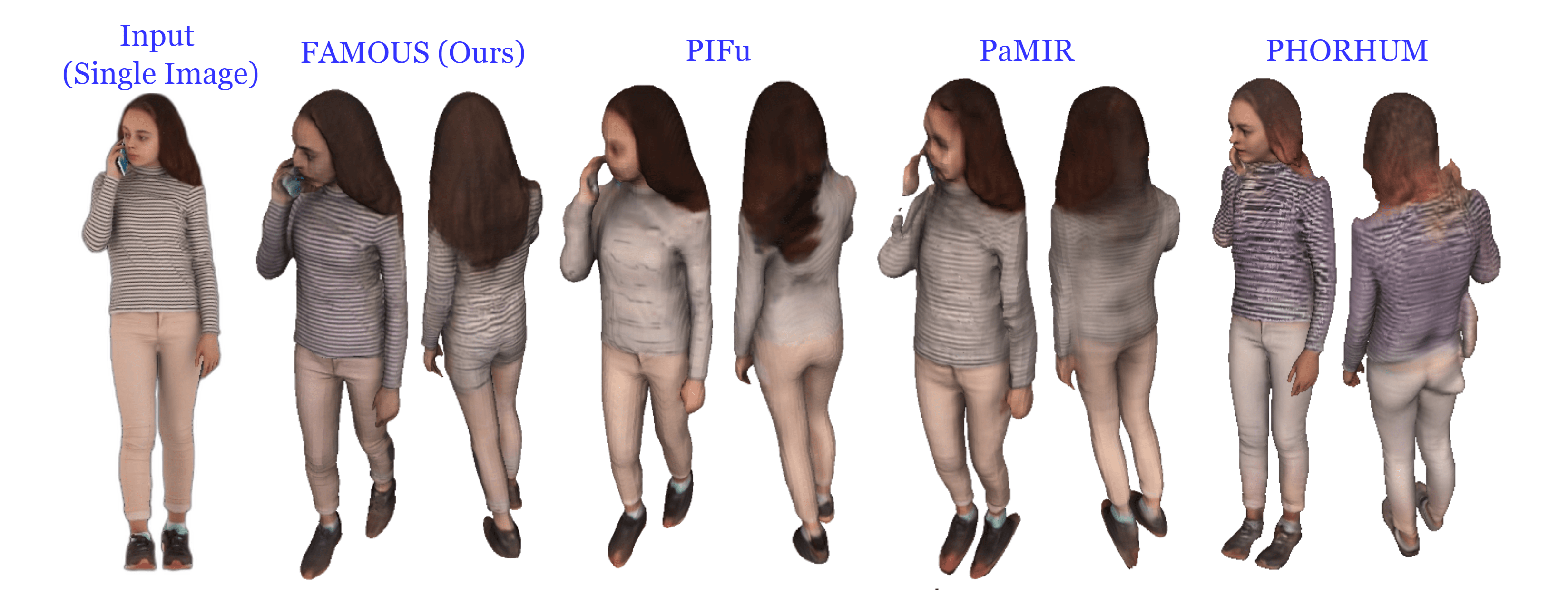}
\vspace{-0.90cm}
\captionof{figure}{
\textbf{Visual comparison} between monocular 3D human digitization methods. Given a single image input, we demonstrate the renderings of textured mesh generated by our approach (FAMOUS) and SOTA pipelines
(\textit{i.e.} predicting both the shape and texture): 
PIFu \cite{pifu}, PaMIR \cite{pamir}, and PHORHUM \cite{phorhum}. Our approach presents significantly improved results in terms of texture and geometry. 
}
\label{fig:diagram}
\end{figure*}

\noindent High-fidelity 3D human digitization (e.g, \cite{pifuhd})  is crucial for a wide range of virtual reality (VR) and augmented reality (AR) applications. This technology is extensively used in industries like gaming and entertainment~\cite{readyplayer}, animation~ \cite{weta-fx}, visual effects, virtual fashion experiences~\cite{walmart-virtual-try-on}, fitness programs~\cite{virtual-fitness}, and virtual training simulations. It  also has significant applications in healthcare, 
anthropology~\cite{virtual-anthropology}, and forensics~\cite{virtual-forensics, virtual-forensics-walkthrough} to name a few. This advanced 3D digitization enables the creation of highly realistic avatars, enhancing the immersive experience in virtual environments by providing detailed and lifelike interactions.

~\\
\noindent Creating such high-precision avatars requires generating both texture and shape in meticulous detail.  Traditionally, this involves collecting data using a 3D scanner and/or a multi-camera setup \cite{multicam}. This data is then processed to create a mesh through algorithms, followed by texture mapping, rigging, and further refinements often involving human artists. However, this semi-manual and multi-step process is not easily scalable for contemporary VR, MR, and AR platforms, which cater to millions of users globally, like the rapidly evolving digital environments of the metaverse.

~\\
Creating an automated method that can handle millions of users is crucial for bringing virtual reality into the mainstream. 
Recent advancements in deep learning have spurred several initiatives towards automation and scalability~\cite{pifu, pifuhd, icon, econ}. However, while the geometric accuracy of these state-of-the-art methods is impressive, the texture quality of unseen areas is still lacking, especially in moderately textured clothing (see Figure \ref{fig:diagram}). 
Some methods even overlook this aspect entirely\cite{icon, econ}.

~\\
Our approach focuses on achieving both realistic texture quality and geometric accuracy, see Figure \ref{fig:diagram}. 
Additionally, we address practical concerns like bandwidth limitations in large-scale VR platforms. 
To tackle this, we propose an approach for 3D human digitization using just a single image per user. Considering the massive user base, ranging from millions to potentially billions, even uploading two images per user would significantly increase bandwidth demands. 

~\\
PIFuHD~\cite{pifuhd} is the first attempt regarding full-resolution monocular 3D human reconstruction that builds on top of the pioneering work of PIFu \cite{pifu} based on implicit function (IF) models. Incorporation of articulated models from the SMPL family \cite{smpl, smplx} to guide the implicit function further improves the geometric reconstruction quality for challenging poses in ICON \cite{icon}. However, this guided reconstruction tends to overfit the articulated model and thus fails to produce realistic results for typical fashion poses (details in the supplementary Section B). Note that among these methods, only PIFu \cite{pifu} predicts both shape and texture (in a coarse resolution), whereas PIFu-HD, ICON, and ECON focus on the reconstruction task only (i.e. no texture). PHORHUM \cite{phorhum} is one of the most recent entry dealing with both human shape and texture based on the albedo surface color and shading information but still fails to extract semantically accurate texture for the occluded region. DIFU \cite{difu} and 2K2K \cite{high} also, effectively reconstructs geometry from high resolution image but DIFU fails to obtain high fidelity textures and 2K2K doesn't focus on texture generation.

~\\
We believe that a key obstacle in achieving satisfactory texture quality in existing literature is the limited diversity of textures in the relatively small pool of available 3D samples~\cite {renderpeople, thuman-2}. Acquiring detailed 3D scans is a time-consuming and costly process. Conversely, there is an abundance of varied cloth textures in 2D images, for example accessible in large-scale 2D fashion datasets~\cite{deep-fashion} and online. 
This contrast points to an opportunity for leveraging these extensive 2D resources to enhance texture quality in 3D models. Drawing from this insight, our paper leverages extensive 2D fashion datasets to enhance the texture quality of 3D models. As will be shown in the experiments section, this approach not only improves texture fidelity but also 
boosts the geometric precision of the models.
~\\
Our approach FAMOUS integrates the rich textural data from 2D datasets into our 3D modeling process through a technique of view synthesis, utilizing pretrained hallucinator~\cite{nted}. We then iteratively refine the hallucinator, focusing on disentangled factors as outlined in our methodology section. This self-supervised process, which we term ``disentangled domain alignment'', effectively aligning to the limited variety found in 3D datasets. Our extensive experiments demonstrate that by merging abundant 2D data with a smaller set of 3D scans in this manner, we can produce 3D models of superior fidelity, both for texture and geometry, compared to existing state-of-the-art methods. 
To our knowledge, this is the first instance of using 2D datasets in conjunction with limited 3D dataset through the lens of hallucinators,  using domain alignment strategy focused on disentanglement factors. 

\noindent Overall, below is the summary of our contributions: 
\vspace{-0.2cm}
\begin{itemize}[leftmargin=*]
\item We propose FAMOUS, a framework that generates a high-fidelity textured mesh given a single RGB image. We harness the rich context available in the form of 2D fashion datasets through the lens of a hallucinator resulting in improved 3D shape and texture, especially for the occluded views. 

\item To this end, we also contribute a large-scale 2D fashion dataset, a derivative from Deep Fashion HD \cite{deep-fashion} but with full body front back image pairs with their COCO-skeleton keypoints annotations. We hope this new dataset will complement future research in the community.

\item We introduce a self-supervised disentangled domain alignment approach to iteratively enable the hallucinator generalize to the distribution of the target 3D dataset. 

\item Finally, we will release our complete codebase along with the dataset splits to facilitate more open-source research in this subdomain.

\end{itemize}




\section{Related Work}
\label{sec:literature}

\subsection{3D Human Digitization}

\noindent\textbf{Multi-view reconstruction.}
 \noindent Earlier work \cite{multiview1, multiview2} in visual hull-based surface reconstruction required images to be captured from multiple viewpoints. The reconstructed avatar needed further postprocessing based on multiview silhouettes to compensate for the additional complexity induced by the topology of the cloth and self-occlusion, in particular. However, the visual hull based approaches fall short on fine-grained reconstruction when the number of input views is limited. On this account, deep volumetric stereo \cite{multiview3} attempts to predict the volumetric occupancy that can capture dynamic clothed humans from highly sparse views. A similar approach is used in \cite{multiview4} that uses a trained autoencoder to generate a deep prior to enable high-end volumetric captures. However, the inherent requirement for a multi-camera setup for these systems poses additional constraints regarding scalability to mass users. \vspace{-1.0ex}

~\\
\noindent \textbf{Single-view reconstruction.} These approaches can be categorized into 
explicit shape-based methods (i.e. voxel grid techniques, template meshes), implicit function-based approaches, and hybrid methods combining both. 
\vspace{-1.0ex}

~\\
\noindent\textbf{1) Explicit shape based approaches.} 
The introduction of SMPL \cite{smpl} made the estimation of pose and shape in 3D human reconstruction tractable. SMPL works based on a linear combination of a small set of body shapes and pose parameters, which allows it to generate to a wide range of realistic body shapes and poses. Octopus \cite{singleview1} takes SMPL further by introducing SMPL-D which learns the SMPL body parameter and additional 3D vertex offsets that model clothing, hair, and details. 
The primary limitation of SMPL and SMPL-D is their fixed topology of the template mesh, which means that these template models are difficult to fit to arbitrary topological changes such as the disappearance of body parts and clothes with substantially different topologies. Other methods use nonparametric depth map \cite{facsimile} or point cloud \cite{pointmodel}. However, scaling up these approaches to model the loose or diverse clothing is nontrivial. 
\vspace{-1.0ex}

%
~\\
\noindent \textbf{2) Implicit function based approaches.} 
This kind of models offer several advantages, such as topology-agnosticism and the ability to represent arbitrary 3D clothed human shapes. PIFu~\cite{pifu} introduces pixel-aligned implicit human shape reconstruction 
-- the first purely implicit function based occupancy and texture predictors for 3D human digitization. PIFuHD \cite{pifuhd} improves the geometric details with multi-resolution network architectures and estimated normal maps. However, both these methods tend to overfit to the body poses in training distribution (i.e. fashion poses)\cite{icon} 
GeoPIFu \cite{geopifu}, Self-Portraits \cite{portraits}, PINA \cite{pina}, and S3 \cite{neuralshape} overcome this limitation by introducing geometric priors to regularize the deep implicit representation. \vspace{-2.0ex}
 
~\\
\noindent \textbf{3) Hybrid appraoches.} PaMIR \cite{pamir}, DeepMultiCap \cite{deepmulticap}, JIFF \cite{jiff}, ARCH \cite{arch}, ARCH++ \cite{archpp} and ICON \cite{icon} attempt to combine both the explicit, parametric body models with continuous, implicit representations. Despite the promising performance of these approaches emphasizing primarily on the modeling innovations, the fundamental limitation regarding the scarcity of detailed 3D scans remains somewhat unexplored. This is evident in case the texture pattern of cloth topology is far from the quite limited training distribution of the 3D scans (Figures \ref{fig:diagram} and \ref{fig:vis}). 
\vspace{-0.2cm}
\subsection{Pose-Guided Person Image Synthesis}
\noindent These approaches synthesize novel views of a person from a reference (source) image and a target pose. Controllable GAN based ones  \cite{personsyn, stylpersonsyn} extract person attributes from various segmentation maps. Dense deformations and flow-based methods \cite{posewithstyle, flow} generate aligned features and estimate appearance flow between references and desired targets. GFLA \cite{spatial} obtains a global flow field and occlusion mask to match the patches from the source image to the target pose. Some other methods \cite{flow, liquidgan} use 3D geometric details that fit SMPL mesh onto 2D images and query the source appearance using the 3D context. PISE \cite{pise} and CASD \cite{crossatt} parse maps to guide the view synthesis in the target pose. CoCosNet \cite{crossdom} applies cross-attention to extract dense correspondences but are limited in lower resolution due to their quadratic memory footprint. Recently, NTED \cite{nted} proposes an efficient attention mechanism 
based on semantic attributes to achieve promising texture quality but lacks 3D consistency.

\section{ Our Approach: FAMOUS}
\label{sec:method}

\begin{figure*}[!htbp]
\centering
\includegraphics[width=0.98\textwidth]{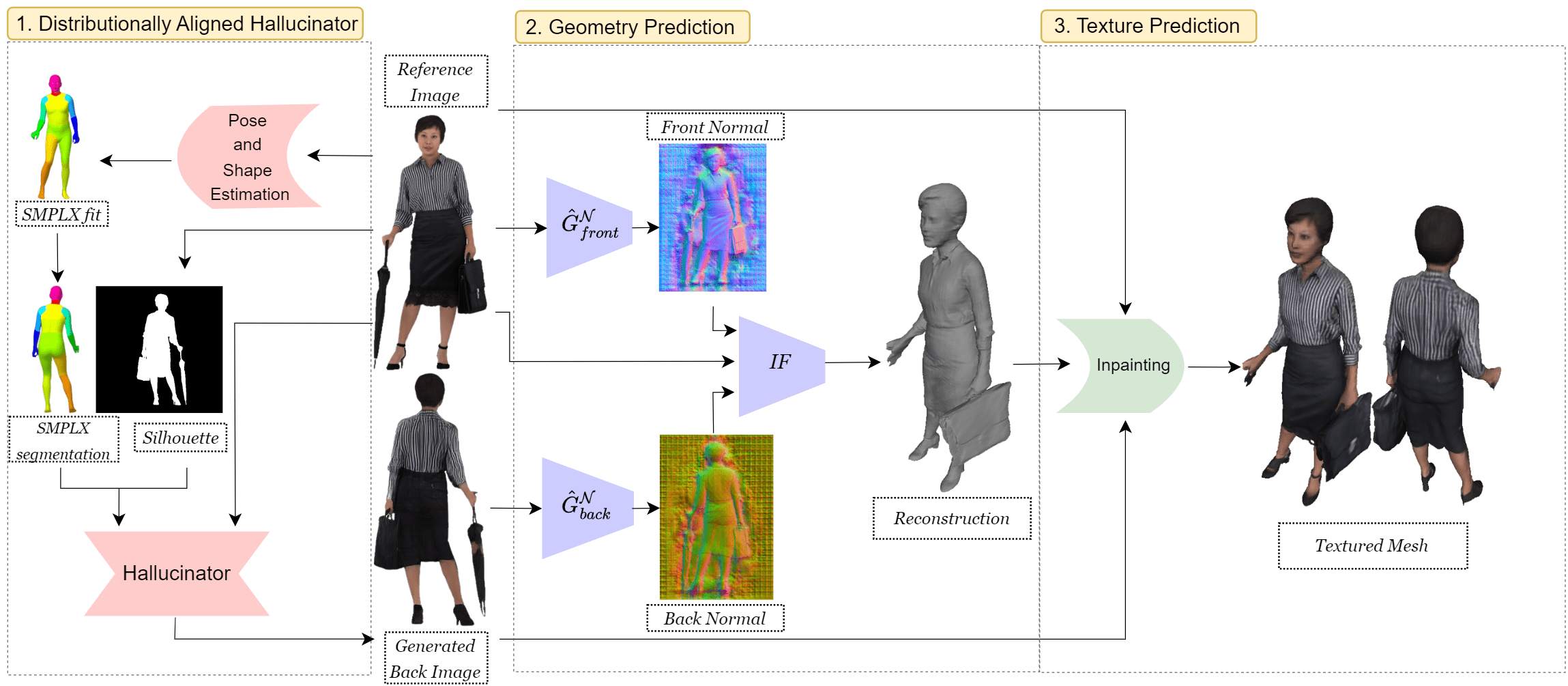}
\vspace{-0.1cm}
\caption{
\textbf{Method overview.} 
Given a single frontal image of a subject, FAMOUS generates a high-quality textured mesh.
Our distributionally aligned hallucinator (Section \ref{subsec:dah}) first predicts the back view based on the reference image, SMPL-X segmentation map and silhouette. Next, the normal map generator $\{(\hat{G}^\mathcal{N}_f, \hat{G}^\mathcal{N}_b)\}$ takes the reference image and generated back image as input and output the respective normal maps. The reference image and the normal estimates are leveraged by the implicit function network (IF) for the geometry prediction (Section \ref{subsec:occ}). Finally, our learnable texture prediction module refines the 3D texture aggregated from the input reference image and predicted back view (Section \ref{subsec:tex}).
}
\label{fig:system}
\vspace{-0.5cm}
\end{figure*}

\noindent
FAMOUS aims to infer high-fidelity 3D shape and texture of a clothed human from a single monocular image. 
Figure~\ref{fig:system} illustrates our complete framework comprising of three steps: (1) Distributionally Aligned Hallucinator (DAH) for generation of the back-view image. (2) Geometry prediction, and (3) texture prediction from the reference image and generated back image. 

\color{black}
\subsection{Distributionally Aligned Hallucinator (DAH)}
\label{subsec:dah}

\noindent A pivotal stage in our process involves generating a back image from the frontal view to provide an additional viewpoint for constructing a 3D model. While this task can be achieved easily using methods trained on 3D scans~\cite{pifu, pamir}, these approaches are typically limited by small training datasets and struggle to generalize well, particularly in term of texture.
In this paper, we propose leveraging large-scale 2D datasets, the source (e.g., Fashion dataset) alongside a limited amount of 3D data, the target (e.g., Render-People dataset~\cite{renderpeople}) to train a robust hallucinator. This hallucinator is designed to generate a back view from a frontal image. Drawing inspiration from recent research~\cite{nted}, we pretrained the hallucinator by augmenting an existing large-scale 2D dataset to account for partial occlusion of the samples (semantic changes) and different body poses (pose changes). Following this augmentation, we conduct simultaneous training and fine-tuning on the target 3D dataset (RenderPeople~\cite{renderpeople}).

\noindent It's worth noting that in our current implementation, we utilize the state-of-the-art sparse attention-based StyleGAN hallucinator~\cite{nted} rather than the most recent diffusion-based approach~\cite{pidm} due to the substantial memory requirements of the latter. 
\subsubsection{Problems with Existing Approaches.}

As mentioned above, SOTA methods for monocular 3D human digitization primarily revolve around training models using 3D scans with limited texture variations. Hence, these approaches fail to achieve good generalization in terms of texture. To overcome this bottleneck, a scalable solution is to leverage the abundant and easily accessible fashion 2D datasets. Directly incorporating the hallucinators \cite{nted, pidm} train on source fashion datasets into our pipeline does not yield satisfactory results for the target 3D dataset. That is, if we train our models in 2D Fashion dataset (e.g., DeepFashion ~\cite{deep-fashion}), and test it in target 3D dataset (e.g., Render people~\cite{renderpeople}), the results are not satisfactory. The same occurs when we fine-tune a pretrained model.  The domain gap between these two datasets leads to performance degradation of the hallucinator on the target data distribution. Domain adaptation methods such as~\cite{mmd, deep-coral} fall short due to the 
insufficient support~\cite{info-theory} of our source and target dataset in terms of pose and texture
variations, respectively (see supplementary Section A for examples and more detailed explanation).  To address this issues, we propose the Disentangled Domain Alignment (DDA) approach.

 \vspace{-0.4cm}
\subsubsection{Disentangled Domain Alignment (DDA)}
\label{subsubsec:dda}

An image can be represented by its style, semantics, pose, and view~\cite{disentaglement}. Our hallucinator\cite{nted} can learn to disentangle these factors from an image to perform a task like view synthesis. We find that the domain gap between the two datasets in the context of the hallucinator depends on these disentanglement factors, so we introduce a factorized alignment approach rooted in this concept. We refer to this approach as DDA. The key steps of our DDA scheme are outlined as follows.

\begin{figure}[ht!]
\centering
\includegraphics[width=0.9\textwidth]{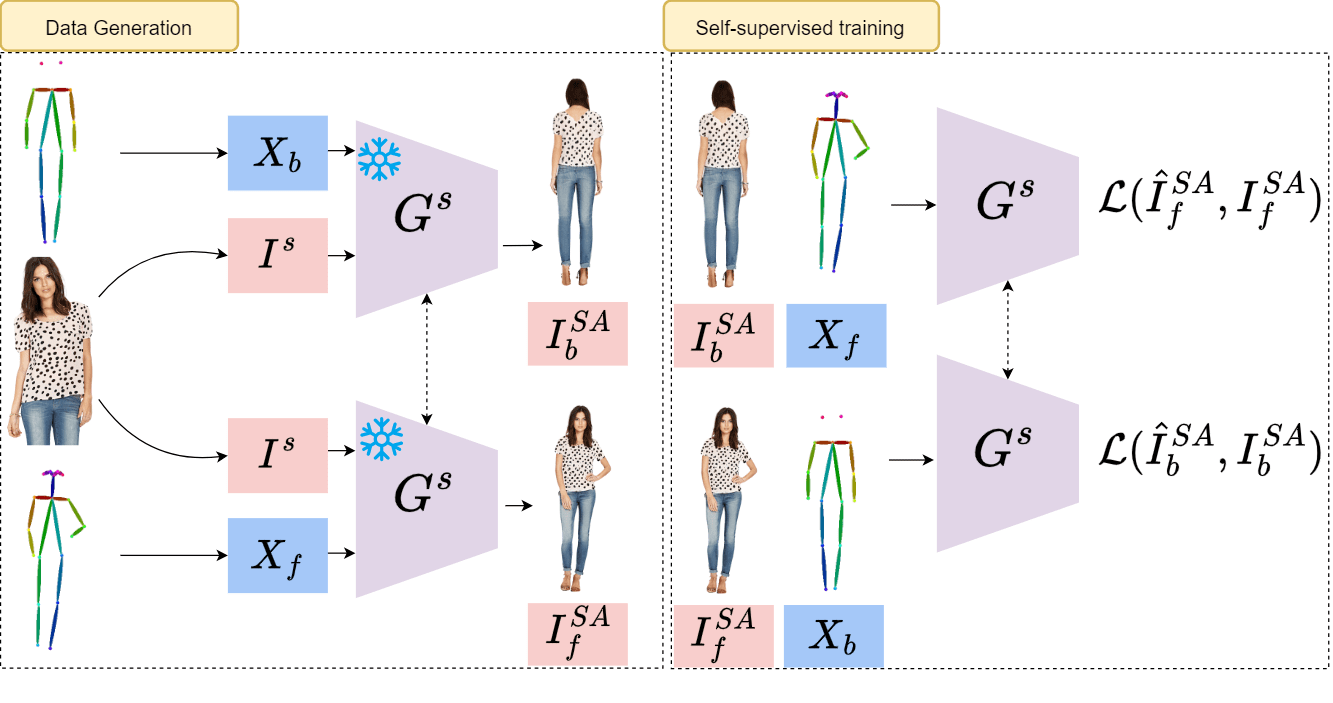}
\vspace{-0.5cm}
\caption{\textbf{Semantic Alignment (SA):}
Given a partial image $I^s$ and full body COCO skeleton ${(X_{f}, X_{b})}$ from our 2D fashion 
dataset (i.e. source), we generate the front/back pairs 
${(I^\mathcal{SA}_f, I^\mathcal{SA}_b )}$
similar to our full body target 3D dataset 
(thus aligning the \textit{semantics}) with a frozen hallucinator ($G^s$). Next, we finetune the hallucinator on these full body pseudo pairs from one another. The hallucinator is initialized with pretrained weights from \cite{nted}, indicated by superscript $s$.  The weight sharing in each stage is indicated by the bidirectional dotted arrows. The default loss function ($\mathcal{L}$) proposed in \cite{nted} is used.
}
\label{fig:dda_sa}
\vspace*{-0.5cm}
\end{figure}

\vspace{-2ex}~\\ 

\noindent \textbf{Semantic Alignment (SA):} The goal of SA is to shift the data distribution learned by the hallucinator from the source to the target distribution based on the semantics. The main difference, in terms of semantics, is the obvious lack of a sufficient number of full body image
pairs covering the front/back view in the source dataset. More specifically, only about $8.5\%$ of the
source fashion dataset \cite{deep-fashion} contains full body pairs and $10\%$ of which encompasses
back views. This is prevalent for the fashion datasets since 
most of these image pairs only highlight a single cloth
(i.e. upper/lower half). To address this limitation, we first generate full body (i.e. target semantics)
image pairs for each sample in the source dataset. For this step, we sample image and guidance with target semantics (i.e. full body COCO key points) from the source dataset and generate full body image pairs with a pretrained hallucinator. Then, we finetune the hallucinator with these generated image pairs in a self-supervised manner, as shown in Figure \ref{fig:dda_sa}. We find this pretraining stage improves the texture prediction, due to the alignment of the hallucinator weights more towards the full body semantic distribution, as will be shown in the experiments section (Table \ref{tab:alignments}). 

\begin{figure}[ht!]
\centering
\includegraphics[width=\textwidth]{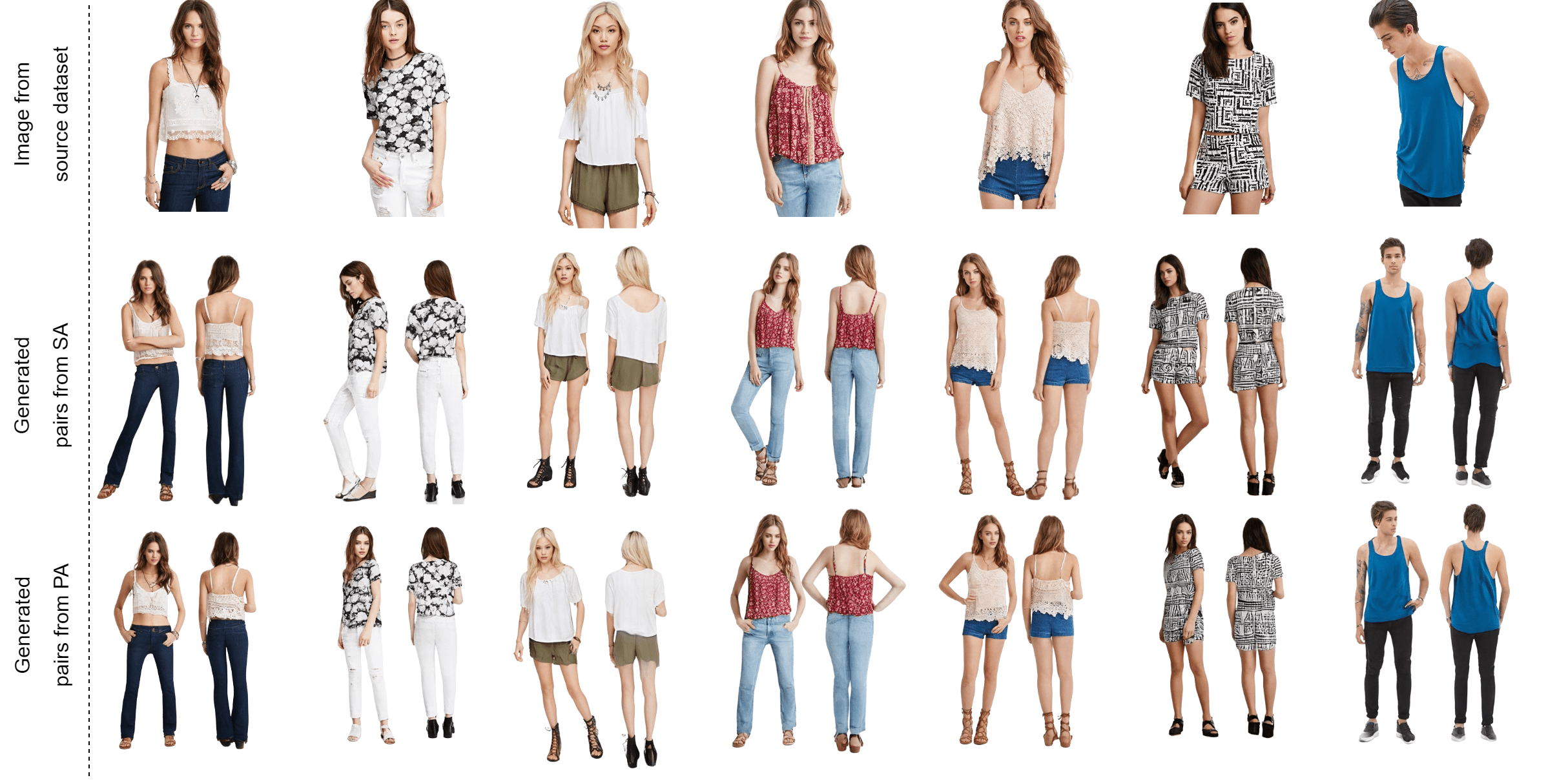}
\vspace{-0.5cm}
\caption{\textbf{Dataset:} The sets of images randomly sampled from the source dataset (top) with their corresponding SA (mid) and PA (bottom) pairs. 
}
\label{fig:data_samples}
\vspace{-0.3cm}
\end{figure}

\noindent To this end, we also release these full body image pairs and their COCO-skeleton keypoints annotations to facilitate future research. A few samples of the generated image pairs are shown in Figure \ref{fig:data_samples}. 
Note that the front/back images do not necessarily mean front/back
canonical fashion poses.

\begin{figure}[ht!]
\centering
\includegraphics[width=0.9\textwidth]{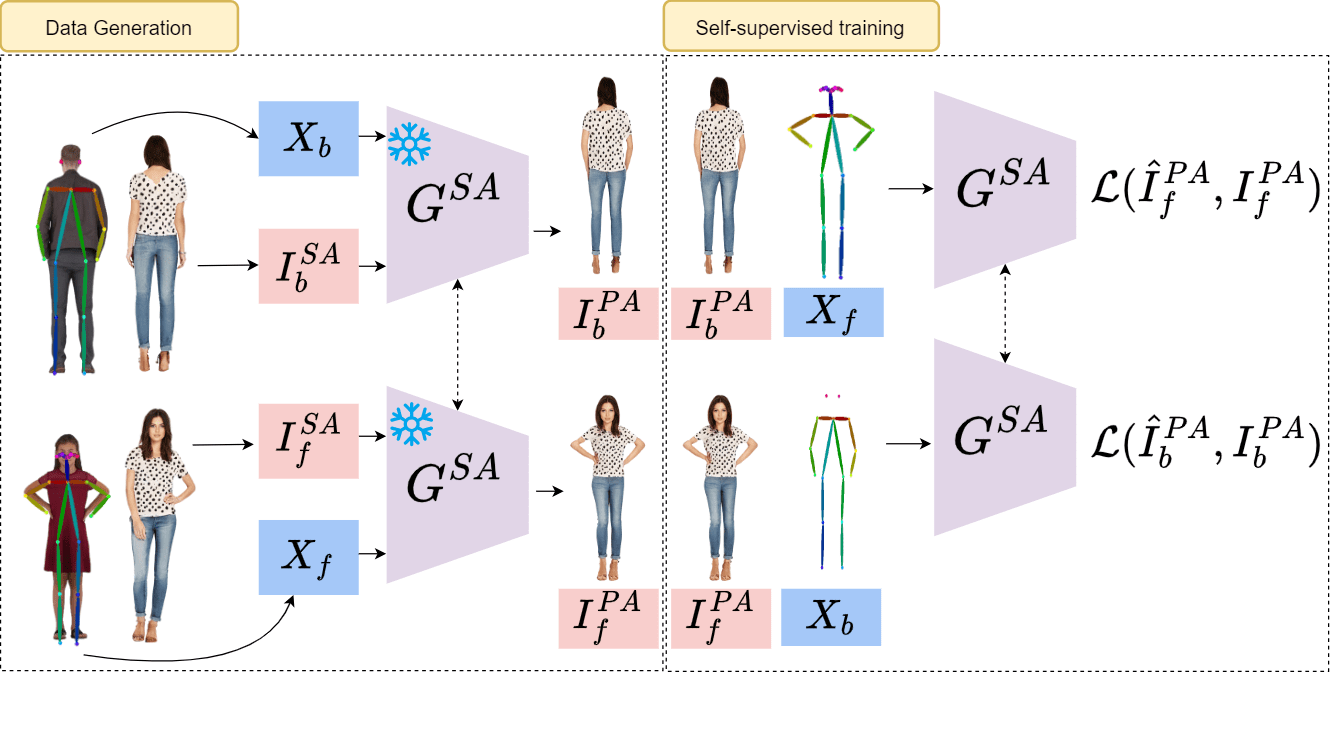}
\vspace{-0.7cm}
\caption{\textbf{Pose Alignment (PA):}Given the subset of the pseudo pairs  
${ ( I^\mathcal{SA}_f, I^\mathcal{SA}_b ) }$
after SA, we generate the front/back pairs 
following the pose from the target distribution 
${ ( I^\mathcal{PA}_f, I^\mathcal{PA}_b ) }$
to finetune the semantically aligned hallucinator further. The hallucinator is initialized with weights obtained after SA.
}
\label{fig:dda_pa}
\vspace{-0.25cm}
\end{figure}

\vspace{-2ex}~\\ 
\textbf{Pose Alignment (PA):} 
Following the semantic alignment, we perform pose alignment to further shift the data distribution learned by our hallucinator based on the target pose distribution. We obtain the pseudo pairs 
$\{ ( I^\mathcal{PA}_f, I^\mathcal{PA}_b ) \}$
from the SA pairs and guidance (with target pose distribution) from the target dataset, as shown in Figure \ref{fig:dda_pa}. 
The sampling for the guidance is based on a discriminator (checkpointed after vanilla hallucinator training \cite{nted}) score thresholding to prevent noisy reconstruction of the PA pairs
(See supplementary Section A for details).
This somewhat conservative threshold allows
us to select a subset with a higher degree of realism that eventually prevents 
degeneracies during further finetuning. Finally, we finetune the semantically aligned hallucinator 
from the last step with these pose-aligned pseudo pairs. Note that we also switch the guidance 
structure (used to represent pose) from COCO keypoints (2D) to SMPL-X segmentation map and silhouette~\cite{smplx}, which we empirically find
in reconstructing the occluded regions more accurately due to its better expressiveness.

\begin{figure}[ht!]
\centering
\includegraphics[width=0.9\textwidth]{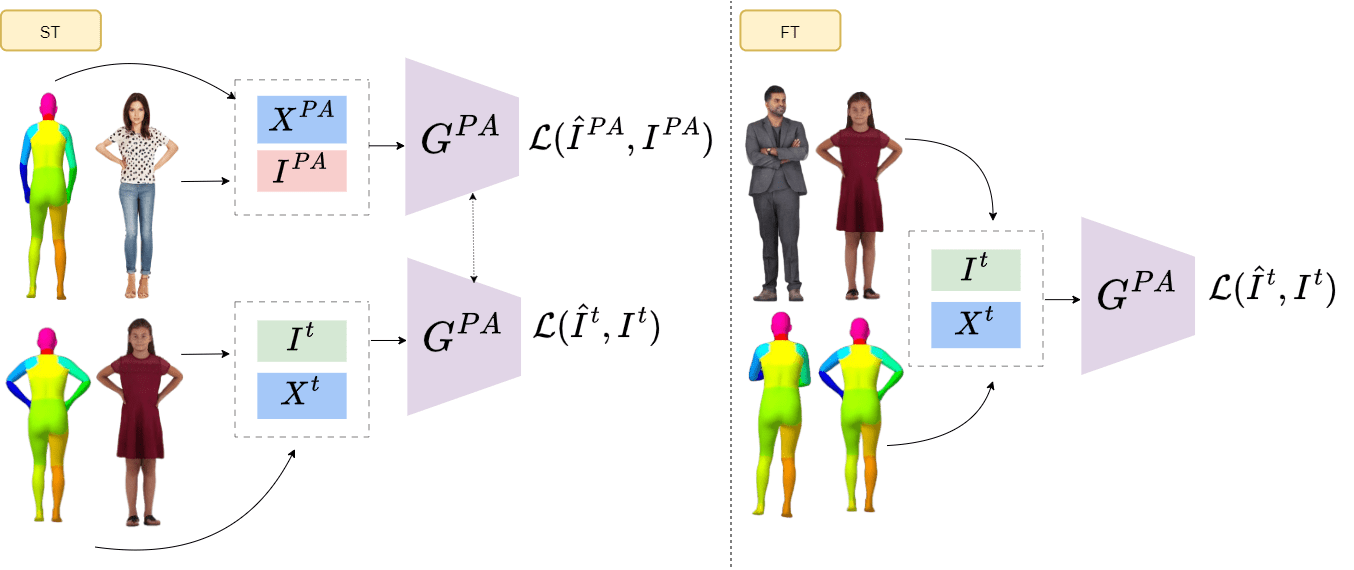}
\vspace{-0.5cm}
\caption{
The updated hallucinator weights after PA is transferred to this stage (indicated by superscript of G). We replace the guidance $X$ from COCO key points to SMPL-X segmentation map for better expressivity.
\textbf{Simultaneous training:} The pseudo pairs after PA, 
$I^\mathcal{PA} $, 
and the 2D renderings of our 
target 3D dataset
$I^t$
are trained together to align the style (i.e. cloth texture and topology) from
both datasets or domains.
\textbf{Finetuning:}
The hallucinator is finetuned directly on the 2D renderings of our 
target 3D dataset.
}
\label{fig:dda}
\vspace{-0.50cm}
\end{figure}

\vspace{-2ex}~\\ 
\textbf{Simultaneous Training (ST) / Finetuning (FT):} At this stage, we obtain 
the pose-aligned pseudo pairs
$\{ ( I^\mathcal{PA}_f, I^\mathcal{PA}_b ) \}$.
This updated subset
represents the source style distribution 
(i.e. texture/clothing patterns) in the target semantic and pose distribution. For ST, this adapted subset
$\{ I^\mathcal{PA} \}$
along with the 2D renderings 
$\{ I^t \}$
from our
target 3D dataset are jointly used to finetune our PA hallucinator for the style alignment (Figure \ref{fig:dda}). In FT, the 2D renderings 
$\{ I^t \}$
are directly used to finetune our PA hallucinator.
In the experiments section, Table~\ref{tab:alignments} shows the ablations of the relative improvements after incorporating
these individual alignment stages in the order prescribed above. During training, we use the GT COCO keypoints, and SMPL-X fits, but for testing, we obtain the SMPL-X fits using off the shelf pose and shape estimation model~\cite{pymafx}.

\color{black}

\subsection{Geometry Prediction}

\noindent This section describes the process for shape generation (using implicit functions) and texture composition. 

\subsubsection{Background on Implicit Functions}

For any point $\mathbf{x} \in \mathbb{R}^3$, the implicit function (IF) methods learn the conditional
probability for occupancy $p(\mathbf{x} | \Pi_\mathbf{x})$, where $\Pi_\mathbf{x}$ 
is the 2D projection of $\mathbf{x}$. 
PIFu \cite{pifu}, the precursor of PIFuHD \cite{pifuhd}, estimates the probability as 
$p(\mathbf{x} | \Pi_\mathbf{x}) = f (\mathcal{F}(\Pi_\mathbf{x}), \Delta)$,
where $\mathcal{F}$ is the pixel aligned feature, and $\Delta$
is the depth estimate of x in camera space. 

~\\
\noindent PIFuHD enhances the vanilla PIFu framework with a (low/high) 
dual-resolution formulation:
\begin{equation}
\label{eq:pifu}
p_{c}(\mathbf{x} | \Pi_\mathbf{x}) = f^{\theta}( 
\color{teal}
f^\lambda (\mathcal{F}_{c}(\Pi_\mathbf{x}), \Delta)    
\color{black}
)
\end{equation}
\noindent where $p_{c}$ is the occupancy field at a coarse-resolution, and $F_{c}$ is the
corresponding pixel-aligned feature.
Also, the implicit function is split into a composition as $f^{\theta}(f^{\lambda}$), where $f^{\lambda}$
jointly contributes to both low/high-resolution occupancy predictions, and $f^{\theta}$ is employed
for low-resolution prediction only. Next, the high-resolution model predicts the final occupancy as: \vspace{-0.1cm}
\begin{equation}
\label{eq:pifuhd}
p(\mathbf{x} | \Pi_\mathbf{x}) = f (\mathcal{F}(\Pi_\mathbf{x}), 
\color{teal}
f^\lambda (\mathcal{F}_{c}(\Pi_\mathbf{x}), \Delta)
\color{black}
)    
\end{equation}
where $\mathcal{F}$ is a separate fine resolution 2D pixel-aligned feature and $f^\lambda(.)$ is the same 
joint embedding from Eq.~\ref{eq:pifu}.

\subsubsection{Occupancy Prediction for Shape Estimation}
\label{subsec:occ}

\vspace{0.1cm}
Our distributionally aligned hallucinator (DAH), as detailed in Section \ref{subsec:dah}, is designed to be versatile and compatible with various shape inference methods. These include purely implicit function (IF) approaches such as PIFu and PIFuHD, as well as hybrid methods that combine implicit functions with explicit articulated model fitting, like SMPL/SMPL-X. Notably, IF approaches generally excel in producing high-quality reconstructions for typical fashion poses, as highlighted in the literature, while hybrid methods tend to shine in handling challenging and acrobatic poses. However, our primary objective is to create lifelike high-quality avatars for AR/VR/MR applications. Once the avatar is initially constructed from a standard pose, further animation of the 3D model becomes feasible, even for complex movements. Consequently, in this paper, we adopt the purely IF-based network, PIFuHD, as the baseline for enhancing DAH. 
\vspace{-1.0ex}


~\\
\noindent Examining Equations \ref{eq:pifu} and \ref{eq:pifuhd}, it is evident that PIFuHD directly estimates depth ($\Delta(\hdots)$). Additionally, pixel aligned features $\mathcal{F}_c$ and $\mathcal{F}$ are leveraged for the occupancy prediction of query 3D points. Our observations indicate that relying solely on the features derived from the back normal map predicted (from front image) by PIFuHD falls short in estimating fine details in the occluded region. To address this limitation, apart from the improved learning of cloth textures from the 2D fashion datasets, we reroute our DAH-predicted back view as input to the back normal estimator (as illustrated in Figure \ref{fig:system}). This modification ultimately results in superior performance for the occluded back region (Please refer to the supplementary Section B for the visualization).


\subsection{Texture prediction}
\label{subsec:tex}

To obtain texture on the reconstructed 3D mesh described earlier, we employ a texture refinement network similar to Tex-PIFu~\cite{pifu}, but with some modifications that 
enhance texture quality.
Firstly, we project the front (reference) and back (synthesized) views onto their corresponding pixel-aligned querry point. This differs from the approach used in PIFu~\cite{pifu}, where masking was not necessary due to the absence of additional synthesized views in the monocular case. Subsequently, our completion network, based on UNet \cite{unet} and MLP, refines the texture at the per-vertex level using an $\mathcal{L}_1$ loss. To achieve more accurate high-frequency texture prediction, the MLP employed for vertex-level feature refinement is equipped with SIREN activation \cite{siren}. For further insights and ablations, please consult the supplementary materials Section C.

\section{Experiments}
\label{sec:experiments}

\noindent \textbf{Datasets: } 
We utilized the RenderPeople dataset \cite{renderpeople} for our evaluation, comprising 950 3D scans. Out of these, 865 scans were allocated for training, while the remaining 85 were reserved for evaluation purposes. To ensure reproducibility, we will make these data splits available alongside our codebase. The 2D renderings derived from these scans were generated at a yaw interval of $10^\circ$. Additionally, we employed PyMAFX \cite{pymafx} to obtain the SMPLX fits and their corresponding segmentation maps. For our 2D fashion dataset, we turned to DeepFashion \cite{deep-fashion}. We specifically used the high-resolution pairs from the original training split in DeepFashion for training, following the approach outlined in our vanilla hallucinator \cite{nted}.






\noindent \textbf{Training and implementation:} Our framework is implemented in PyTorch. The disentangled domain alignement scheme is implemented on top of the vanilla 2D hallucinator codebase \cite{nted}. Moreover, the occupancy and texture completion subsystems are built on the official releases containing partial implementations only \cite{pifuhd}. Unlike these official implementations, we will make our complete source code publicly available to facilitate further research into this direction. For DDA and occupancy prediction, we follow the same hyperparameter settings from the respective literature \cite{nted, pifu, pifuhd}. For texture completion, however, we use the Adam optimizer with the learning rate of $1e^{-4}$ to train for $20$ epochs. For the 3D dataset, we generate 2D renderings at $10^\circ$ intervals. All the models are trained on a dedicated \texttt{8-GPU NVIDIA RTX A4000} server.

\begin{table}[!t]
\begin{center}
\caption{Quantitative comparison of texture and shape on RenderPeople \cite{renderpeople} test set.}
\vspace{-0.5cm}
\label{tab:sota}
\begin{adjustbox}{max width=\textwidth}
\renewcommand{\arraystretch}{1.3}
\begin{threeparttable}
\begin{tabular}{l|cccc|ccc}
\hline
\multicolumn{1}{c|}{\multirow{2}{*}{Methods}} & \multicolumn{4}{c|}{Texture} & \multicolumn{3}{c}{Shape} \\ \cline{2-8} 
\multicolumn{1}{c|}{} & FID ($\downarrow$) & IS ($\uparrow$) & LPIPS ($\downarrow$) & SSIM ($\uparrow$) & Chamfer ($\downarrow$)
& P2S ($\downarrow$) & Nml MSE ($\downarrow$) \\ \hline
PIFu  \cite{pifu} & 67.6852 & 3.5190 & 0.1551 & 0.9101 & 1.6631 & 1.4427 & 0.0554\\
PAMIR \cite{pamir} & 70.8481 & 3.6684 &  0.1542  &  0.8963  & 2.6157 & 2.0091 & 0.0738\\
PIFuHD \cite{pifuhd} & - & - & - & - & 1.5709 & 1.4025 & 0.0526 \\
ICON \cite{icon} & - & - & - & - & 2.4936 & 2.0389  & 0.0680 \\
ECON \cite{econ} & - & - & - & - & 2.7133 & 2.4414 &0.0742 \\
FAMOUS (Ours) & \textbf{55.0711} & \textbf{3.8655} & \textbf{0.1425} & \textbf{0.9140} & \textbf{1.4102} & \textbf{1.2718} & \textbf{0.0371}  \\
\hline
\end{tabular}
\begin{tablenotes}
\item []
\end{tablenotes}
\end{threeparttable}
\end{adjustbox}
\end{center}
\vspace{-0.5cm}
\end{table}

\vspace{-0.5cm}
\subsection{Evaluation}

\noindent \textbf{SOTA comparison:} Table \ref{tab:sota} shows the comparison with the SOTA methods both in terms of texture and shape. The texture scores are not reported for the methods focusing only on shape (i.e. PIFuHD \cite{pifuhd}, ICON \cite{icon}, and ECON \cite{econ}). 
\vspace{-0.5cm}

\begin{figure}[!t]
\centering
\includegraphics[width=0.96\textwidth]{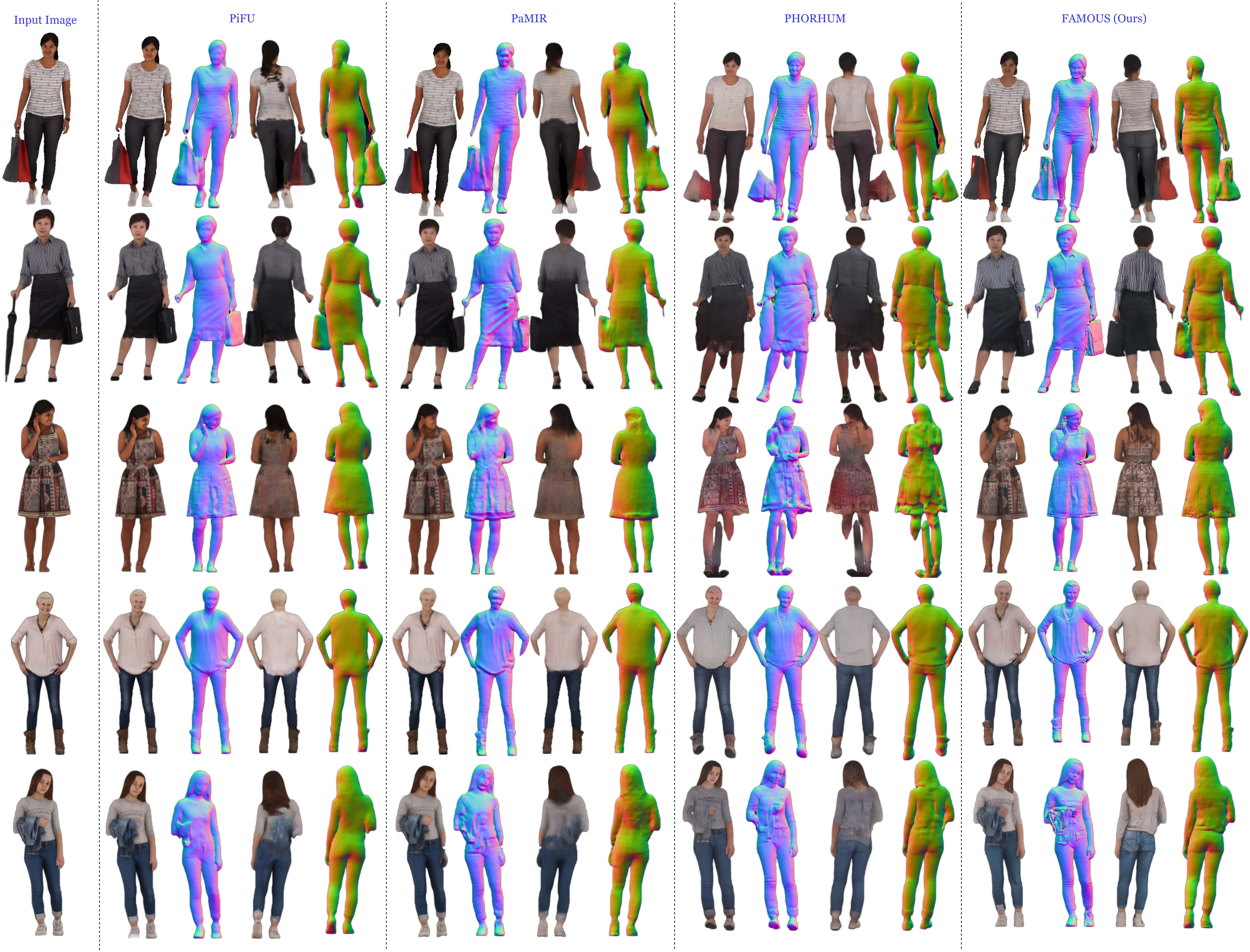}
\vspace*{-0.3cm}
\caption{Qualitative comparison of the SOTA pipelines predicting both shape and texture with ours
on the RenderPeople test set for the images shown on the left.
From left to right: Input image, PIFu \cite{pifu}, PaMIR \cite{pamir}, PHORHUM \cite{phorhum}, and FAMOUS (ours).
For each of these methods, we provide 4 visualizations -- front orthographic view (2D rendering), front surface normals,
back orthographic view (2D rendering), and back surface normals in order. FAMOUS visually outperforms these SOTA approaches in predicting both the occluded texture and surface delicacies.
}
\label{fig:vis}
\end{figure}

~\\
In terms of texture quality, our method, FAMOUS, outperforms its counterparts in $3$ out of $4$ metrics, achieving approximately $19\%$ and $6\%$ relative improvement in FID and IS, respectively, over the second-best alternative. For SSIM, PIFu \cite{pifu} exhibits similar performance compared to ours. It is important to note that SSIM places greater emphasis on smoother predictions with reduced high-frequency variations, and this aligns with the characteristics of PIFu's predictions, as evidenced in Figure \ref{fig:vis}. In the figure, the sharp, localized variations in both normals and textures from PIFu appear as a subdued version of the ground truth, whereas FAMOUS preserves the intricate details that are captured by deep perceptual metrics, such as FID and IS. \vspace{-1.5ex}

~\\
Regarding shape reconstruction, our method outperforms the SOTA methods across the board, with PIFuHD \cite{pifuhd} as the second-best contender (Table \ref{tab:sota}). Note that the recent shape-only approaches, i.e. ICON \cite{icon} and ECON \cite{econ}, (with the publicly available codebases open-sourced by the respective authors), perform worse than PIFuHD. Based on further qualitative investigation, we find these SMPLX model based methods excel others for the more difficult acrobatic poses. Such extreme poses are not present in Fashion images, and PIFuHD is preferred for those cases\cite{econ}. On a related note, RenderPeople \cite{renderpeople}, our target 3D dataset, contains mostly casual poses -- somewhat similar to the canonical case, and so, more appropriate for our evaluation than the extreme cases. Nonetheless, we include them for completeness. Please refer to the supplementary Section B for additional visualizations in this regard.


\begin{table}[!t]
\begin{center}
\vspace{-0.3cm}
\caption{Ablation study of the hallucinator on Render People.}
\vspace{-1.5ex}
\label{tab:alignments}
\begin{adjustbox}{max width=0.58\textwidth}
\renewcommand{\arraystretch}{1.3}
\begin{threeparttable}
\begin{tabular}{l|cccc}
\hline
Experiments & FID($\downarrow$) & LPIPS($\downarrow$) & SSIM ($\uparrow$)& IS ($\uparrow$)\\ \hline
FT & 77.1910 & 0.1007 & 0.8645 & 2.7750\\
FT+SA & 73.0653 & 0.0935 & 0.8678 & 2.7222\\
FT+SA+PA & 67.7944 & 0.0824 & \textbf{0.8929} & 2.6282\\
\hdashline
ST & 71.0900 & 0.0935 & 0.8678 & 2.7228\\
ST+SA & 77.2460 & 0.0927 & 0.8683 & 2.5015\\
ST+SA+PA & \textbf{66.8911} & \textbf{0.0822} & 0.8899 & \textbf{3.2810} \\
\hline
\end{tabular}
\begin{tablenotes}
\item [] \hspace{-0.4cm} \small FT $\equiv$ Finetuning; \hspace{0.2cm} ST $\equiv$ Simultaneous Training 
\item [] \hspace{-0.4cm} \small SA $\equiv$ Semantic Alignment;  PA $\equiv$ Pose Alignment
\end{tablenotes}
\end{threeparttable}
\end{adjustbox}
\end{center}
\vspace{-0.50cm}
\end{table}

\subsection{Ablation Studies}

\noindent \textbf{Disentangled domain alignment (DDA):} Table \ref{tab:alignments}
shows progressive improvements for the back (occluded) view prediction with our DDA alignments on the pretrained hallucinator (Section \ref{subsec:dah}). DDA improves the prediction quality for both finetuning (FT) on the 2D renderings of our 3D dataset and simultaneously training (ST) with both \textbf{aligned} 2D fashion images (source) and the 2D renderings (target). \vspace{-0.5ex}

~\\
\noindent For finetuning, one thing to note is that IS keeps degrading as we keep doing more alignment for finetuning (FT $\rightarrow$ FT+SA $\rightarrow$ FT+SA+PA) whereas the other $3$ metrics gradually improve. This is because the distribution of the generated samples in each stage of alignment gets closer to the distribution of 2D renderings on the test set (gradually lower FID). However, the 3D dataset lacks sufficient number of samples with diverse textures, and so, finetuning with just the corresponding 2D renderings lead to an apparent loss of realism reflected by IS degradation. \vspace{-1.0ex}

~\\
\noindent For simultaneous training (ST), FID and IS degrade while LPIPS and SSIM improve slightly after incorporating the semantic alignment (SA). This is because the SA images from the source (2D fashion dataset) contains predictions with variable amount of realism -- some of which are failed cases. Simultaneously tuning the hallucinator with this unfiltered set of generated samples alongside 2D renderings lead to poor convergence which is the reason behind the high perceptual losses for ST+SA in Table \ref{tab:alignments}. However, the additional discriminator scoring in the beginning of pose alignment helps prevent such convergence issues, thus leading to significant improvements in the final stage (ST+SA \textit{vs.} ST+SA+PA). Also, simultaneously training with the aligned 2D fashion images and the 2D renderings equivalent to Style Alignment (TA) as described in Section \ref{subsec:dah}. Thus, the improvement of (ST+SA+PA) over (FT+SA+PA) in Table \ref{tab:alignments} shows the effectiveness of our complete disentangled alignment. Lastly, SSIM for FT is slightly better than ST due to the potential overfitting of the sufficient statistics on the 2D renderings in case of FT where the support of ST \cite{info-theory} will likely generalize better.

\section{Conclusion and Future Work}
\label{sec:conclusion}

\noindent
This paper introduces a complete framework for 3D human digitization from one image, emphasizing the generation of high-fidelity textures. In contrast to SOTA methods, we propose a new approach that harnesses the wealth of 2D datasets to predict arbitrary texture patterns, addressing the scarcity of similar 3D datasets. To achieve this, we leverage recent advancements in 2D hallucinators, incorporating a gradual domain alignment strategy based on disentangled factors. This integration of information from 2D datasets leads to a significant increase in texture prediction accuracy while also improving the accuracy of shape inference. To the best of our knowledge, this represents the first work to primarily focus on enhancing the quality of arbitrary textures for 3D human digitization by utilizing large-scale 2D datasets.  
\vspace{-1.0ex}

~\\
In terms of limitations and future directions, both 3D texture and shape predictions face challenges when dealing with rare clothing types that are not present in either the 2D/3D datasets. Additionally, texture prediction may encounter difficulties with rare or highly acrobatic poses that fall outside the scope of the training distribution.

\newpage

%
%
\bibliographystyle{splncs04}
\bibliography{
    bibliographies/misc,
    bibliographies/human3d, 
    bibliographies/gan2d,
    bibliographies/datasets,
    bibliographies/da
}

\begin{thebibliography}{10}
\providecommand{\url}[1]{\texttt{#1}}
\providecommand{\urlprefix}{URL }
\providecommand{\doi}[1]{https://doi.org/#1}

\bibitem{virtual-forensics}
{Forensic Science Professor Brings Her Innovative VR Tech to Roger Williams University}. \url{https://www.rwu.edu/news/news-archive/forensic-science-professor-brings-her-innovative-vr-tech-rwu}

\bibitem{virtual-fitness}
{IGOODI Official}. \url{https://www.linkedin.com/pulse/virtual-fitness-training-future-realistic-avatars-igoodi-official}

\bibitem{readyplayer}
{Readyplayer}. \url{https://readyplayer.me}

\bibitem{renderpeople}
{RenderPeople}. \url{https://renderpeople.com/3d-people}

\bibitem{virtual-anthropology}
{Virtual Fieldwork: Using Virtual Reality (VR) in Anthropology}. \url{https://www.unf.edu/dhi/projects/current/virtual-fieldwork-using-virtual-reality-in-anthropology.html}

\bibitem{walmart-virtual-try-on}
{Walmart Virtual Try-On}. \url{https://www.walmart.com/cp/virtual-try-on/4879497}

\bibitem{weta-fx}
{Weta FX}. \url{https://www.wetafx.co.nz/films}

\bibitem{posewithstyle}
AlBahar, B., Lu, J., Yang, J., Shu, Z., Shechtman, E., Huang, J.B.: Pose with {S}tyle: {D}etail-preserving pose-guided image synthesis with conditional stylegan. ACM ToG  (2021)

\bibitem{singleview1}
Alldieck, T., Magnor, M., Bhatnagar, B.L., Theobalt, C., Pons-Moll, G.: Learning to reconstruct people in clothing from a single rgb camera. In: CVPR (2019)

\bibitem{phorhum}
Alldieck, T., Zanfir, M., Sminchisescu, C.: Photorealistic monocular 3d reconstruction of humans wearing clothing. In: CVPR (2022)

\bibitem{pidm}
Bhunia, A.K., Khan, S., Cholakkal, H., Anwer, R.M., Laaksonen, J., Shah, M., Khan, F.S.: Person image synthesis via denoising diffusion model. In: CVPR (2023)

\bibitem{jiff}
Cao, Y., Chen, G., Han, K., Yang, W., Wong, K.Y.K.: {JIFF}: Jointly-aligned implicit face function for high quality single view clothed human reconstruction. In: CVPR (2022)

\bibitem{info-theory}
Cover, T.M., Thomas, J.A.: Elements of Information Theory. Wiley-Interscience, USA (2006)

\bibitem{pina}
Dong, Z., Guo, C., Song, J., Chen, X., Geiger, A., Hilliges, O.: {PINA}: Learning a personalized implicit neural avatar from a single rgb-d video sequence. In: CVPR (2022)

\bibitem{multiview4}
Gilbert, A., Volino, M., Collomosse, J., Hilton, A.: Volumetric performance capture from minimal camera viewpoints. In: ECCV (2018)

\bibitem{multicam}
Guo, K., Lincoln, P., Davidson, P., Busch, J., Yu, X., Whalen, M., Harvey, G., Orts-Escolano, S., Pandey, R., Dourgarian, J., Tang, D., Tkach, A., Kowdle, A., Cooper, E., Dou, M., Fanello, S., Fyffe, G., Rhemann, C., Taylor, J., Debevec, P., Izadi, S.: The relightables: Volumetric performance capture of humans with realistic relighting. ACM ToG  (2019)

\bibitem{high}
Han, S.H., Park, M.G., Yoon, J.H., Kang, J.M., Park, Y.J., Jeon, H.G.: High-fidelity 3d human digitization from single 2k resolution images. In: Proceedings of the IEEE/CVF Conference on Computer Vision and Pattern Recognition (CVPR) (2023)

\bibitem{geopifu}
He, T., Collomosse, J., Jin, H., Soatto, S.: {Geo-PIFu}: Geometry and pixel aligned implicit functions for single-view human reconstruction. In: NeurIPS (2020)

\bibitem{archpp}
He, T., Xu, Y., Saito, S., Soatto, S., Tung, T.: {ARCH++}: Animation-ready clothed human reconstruction revisited. In: ICCV (2021)

\bibitem{multiview3}
Huang, Z., Li, T., Chen, W., Zhao, Y., Xing, J., LeGendre, C., Luo, L., Ma, C., Li, H.: Deep volumetric video from very sparse multi-view performance capture. In: ECCV (2018)

\bibitem{arch}
Huang, Z., Xu, Y., Lassner, C., Li, H., Tung, T.: {ARCH}: Animatable reconstruction of clothed humans. In: CVPR (2020)

\bibitem{flow}
Li, Y., Huang, C., Loy, C.C.: Dense intrinsic appearance flow for human pose transfer. In: CVPR (2019)

\bibitem{portraits}
Li, Z., Yu, T., Zheng, Z., Liu, Y.: Robust and accurate 3d self-portraits in seconds. IEEE TPAMI  (2022)

\bibitem{liquidgan}
Liu, W., Piao, Z., Min, J., Luo, W., Ma, L., Gao, S.: Liquid warping gan: A unified framework for human motion imitation, appearance transfer and novel view synthesis. In: ICCV (2019)

\bibitem{deep-fashion}
Liu, Z., Luo, P., Qiu, S., Wang, X., Tang, X.: Deepfashion: Powering robust clothes recognition and retrieval with rich annotations. In: CVPR (2016)

\bibitem{smpl}
Loper, M., Mahmood, N., Romero, J., Pons-Moll, G., Black, M.J.: {SMPL}: A skinned multi-person linear model. ACM Trans. Graph.  \textbf{34}(6) (2015)

\bibitem{personsyn}
Men, Y., Mao, Y., Jiang, Y., Ma, W.Y., Lian, Z.: Controllable person image synthesis with attribute-decomposed gan. In: CVPR (2020)

\bibitem{smplx}
Pavlakos, G., Choutas, V., Ghorbani, N., Bolkart, T., Osman, A.A.A., Tzionas, D., Black, M.J.: Expressive body capture: {3D} hands, face, and body from a single image. In: CVPR (2019)

\bibitem{disentaglement}
Prabhudesai, M., Lal, S., Patil, D., Tung, H.Y., Harley, A.W., Fragkiadaki, K.: Disentangling 3d prototypical networks for few-shot concept learning. In: ICLR (2021)

\bibitem{nted}
Ren, Y., Fan, X., Li, G., Liu, S., Li, T.H.: Neural texture extraction and distribution for controllable person image synthesis. In: CVPR (2022)

\bibitem{spatial}
Ren, Y., Yu, X., Chen, J., Li, T.H., Li, G.: Deep image spatial transformation for person image generation. In: CVPR (2020)

\bibitem{unet}
Ronneberger, O., Fischer, P., Brox, T.: U-net: Convolutional networks for biomedical image segmentation. In: MICCAI (2015)

\bibitem{pifu}
Saito, S., Huang, Z., Natsume, R., Morishima, S., Kanazawa, A., Li, H.: {PIFu}: Pixel-aligned implicit function for high-resolution clothed human digitization. In: ICCV (2019)

\bibitem{pifuhd}
Saito, S., Simon, T., Saragih, J., Joo, H.: {PIFuHD}: Multi-level pixel-aligned implicit function for high-resolution 3d human digitization. In: CVPR (2020)

\bibitem{stylpersonsyn}
Sarkar, K., Golyanik, V., Liu, L., Theobalt, C.: Style and pose control for image synthesis of humans from a single monocular view (2021)

\bibitem{virtual-forensics-walkthrough}
Sieberth, T., Dobay, A., Affolter, R., Ebert, L.C.: Applying virtual reality in forensics – a virtual scene walkthrough. Forensic Science, Medicine and Pathology  (2019)

\bibitem{siren}
Sitzmann, V., Martel, J., Bergman, A., Lindell, D., Wetzstein, G.: Implicit neural representations with periodic activation functions. In: NeurIPS (2020)

\bibitem{facsimile}
Smith, D., Loper, M., Hu, X., Mavroidis, P., Romero, J.: Facsimile: Fast and accurate scans from an image in less than a second. In: ICCV (2019)

\bibitem{difu}
Song, D.Y., , Lee, H., Seo, J., Cho, D.: Difu: Depth-guided implicit function for clothed human reconstruction (2023)

\bibitem{deep-coral}
Sun, B., Saenko, K.: Deep coral: Correlation alignment for deep domain adaptation. In: ECCV Workshops (2016)

\bibitem{mmd}
Tzeng, E., Hoffman, J., Zhang, N., Saenko, K., Darrell, T.: Deep domain confusion: Maximizing for domain invariance. CoRR  \textbf{abs/1412.3474} (2014)

\bibitem{multiview1}
Vlasic, D., Baran, I., Matusik, W., Popovi\'{c}, J.: Articulated mesh animation from multi-view silhouettes. ACM Trans. Graph.  (2008)

\bibitem{econ}
Xiu, Y., Yang, J., Cao, X., Tzionas, D., Black, M.J.: {ECON}: Explicit clothed humans optimized via normal integration. In: CVPR (2023)

\bibitem{icon}
Xiu, Y., Yang, J., Tzionas, D., Black, M.J.: {ICON}: {I}mplicit {C}lothed humans {O}btained from {N}ormals. In: CVPR (2022)

\bibitem{neuralshape}
Yang, Z., Wang, S., Manivasagam, S., Huang, Z., Ma, W.C., Yan, X., Yumer, E., Urtasun, R.: S3: Neural shape, skeleton, and skinning fields for 3d human modeling. In: CVPR (2021)

\bibitem{thuman-2}
Yu, T., Zheng, Z., Guo, K., Liu, P., Dai, Q., Liu, Y.: Function4d: Real-time human volumetric capture from very sparse consumer rgbd sensors. In: CVPR (2021)

\bibitem{pointmodel}
Zakharkin, I., Mazur, K., Grigorev, A., Lempitsky, V.: Point-based modeling of human clothing. In: ICCV (2021)

\bibitem{pymafx}
Zhang, H., Tian, Y., Zhang, Y., Li, M., An, L., Sun, Z., Liu, Y.: {PyMAF-X}: Towards well-aligned full-body model regression from monocular images. IEEE TPAMI  (2023)

\bibitem{pise}
Zhang, J., Li, K., Lai, Y.K., Yang, J.: Pise: Person image synthesis and editing with decoupled gan. In: CVPR (2021)

\bibitem{crossdom}
Zhang, P., Zhang, B., Chen, D., Yuan, L., Wen, F.: Cross-domain correspondence learning for exemplar-based image translation. In: CVPR (2020)

\bibitem{deepmulticap}
Zheng, Y., Shao, R., Zhang, Y., Yu, T., Zheng, Z., Dai, Q., Liu, Y.: Deepmulticap: Performance capture of multiple characters using sparse multiview cameras. In: ICCV (2021)

\bibitem{pamir}
Zheng, Z., Yu, T., Liu, Y., Dai, Q.: {PaMIR}: Parametric model-conditioned implicit representation for image-based human reconstruction. IEEE TPAMI  \textbf{44}(6) (2022)

\bibitem{crossatt}
Zhou, X., Yin, M., Chen, X., Sun, L., Gao, C., Li, Q.: Cross attention based style distribution for controllable person image synthesis. In: ECCV (2022)

\bibitem{multiview2}
Zuo, X., Du, C., Wang, S., Zheng, J., Yang, R.: Interactive visual hull refinement for specular and transparent object surface reconstruction. In: ICCV (2015)

\end{thebibliography}


\begin{thebibliography}{10}
\providecommand{\url}[1]{\texttt{#1}}
\providecommand{\urlprefix}{URL }
\providecommand{\doi}[1]{https://doi.org/#1}

\bibitem{phorhum}
Alldieck, T., Zanfir, M., Sminchisescu, C.: Photorealistic monocular 3d reconstruction of humans wearing clothing. In: CVPR (2022)

\bibitem{disentaglement}
Prabhudesai, M., Lal, S., Patil, D., Tung, H.Y., Harley, A.W., Fragkiadaki, K.: Disentangling 3d prototypical networks for few-shot concept learning. In: ICLR (2021)

\bibitem{nted}
Ren, Y., Fan, X., Li, G., Liu, S., Li, T.H.: Neural texture extraction and distribution for controllable person image synthesis. In: CVPR (2022)

\bibitem{pifu}
Saito, S., Huang, Z., Natsume, R., Morishima, S., Kanazawa, A., Li, H.: {PIFu}: Pixel-aligned implicit function for high-resolution clothed human digitization. In: ICCV (2019)

\bibitem{pifuhd}
Saito, S., Simon, T., Saragih, J., Joo, H.: {PIFuHD}: Multi-level pixel-aligned implicit function for high-resolution 3d human digitization. In: CVPR (2020)

\bibitem{siren}
Sitzmann, V., Martel, J., Bergman, A., Lindell, D., Wetzstein, G.: Implicit neural representations with periodic activation functions. In: NeurIPS (2020)

\bibitem{deep-coral}
Sun, B., Saenko, K.: Deep coral: Correlation alignment for deep domain adaptation. In: ECCV Workshops (2016)

\bibitem{mmd}
Tzeng, E., Hoffman, J., Zhang, N., Saenko, K., Darrell, T.: Deep domain confusion: Maximizing for domain invariance. CoRR  \textbf{abs/1412.3474} (2014)

\bibitem{econ}
Xiu, Y., Yang, J., Cao, X., Tzionas, D., Black, M.J.: {ECON}: Explicit clothed humans optimized via normal integration. In: CVPR (2023)

\bibitem{icon}
Xiu, Y., Yang, J., Tzionas, D., Black, M.J.: {ICON}: {I}mplicit {C}lothed humans {O}btained from {N}ormals. In: CVPR (2022)

\bibitem{pymafx}
Zhang, H., Tian, Y., Zhang, Y., Li, M., An, L., Sun, Z., Liu, Y.: {PyMAF-X}: Towards well-aligned full-body model regression from monocular images. IEEE TPAMI  (2023)

\bibitem{pamir}
Zheng, Z., Yu, T., Liu, Y., Dai, Q.: {PaMIR}: Parametric model-conditioned implicit representation for image-based human reconstruction. IEEE TPAMI  \textbf{44}(6) (2022)

\end{thebibliography}
\end{document}


\title{FAMOUS: High-\textcolor{blue!50!white}{F}idelity Monocul\textcolor{blue!50!white}{a}r \\ 
3D Hu\textcolor{blue!50!white}{m}an Digitizati\textcolor{blue!50!white}{o}n \textcolor{blue!50!white}{U}sing View \textcolor{blue!50!white}{S}ynthesis}

\titlerunning{3D Human Digitization}

\author{
}

\institute{}

\maketitle

\begin{figure}
\setlength{\linewidth}{\textwidth}
\setlength{\hsize}{\textwidth}
\centering
\includegraphics[height = 0.85\textwidth]{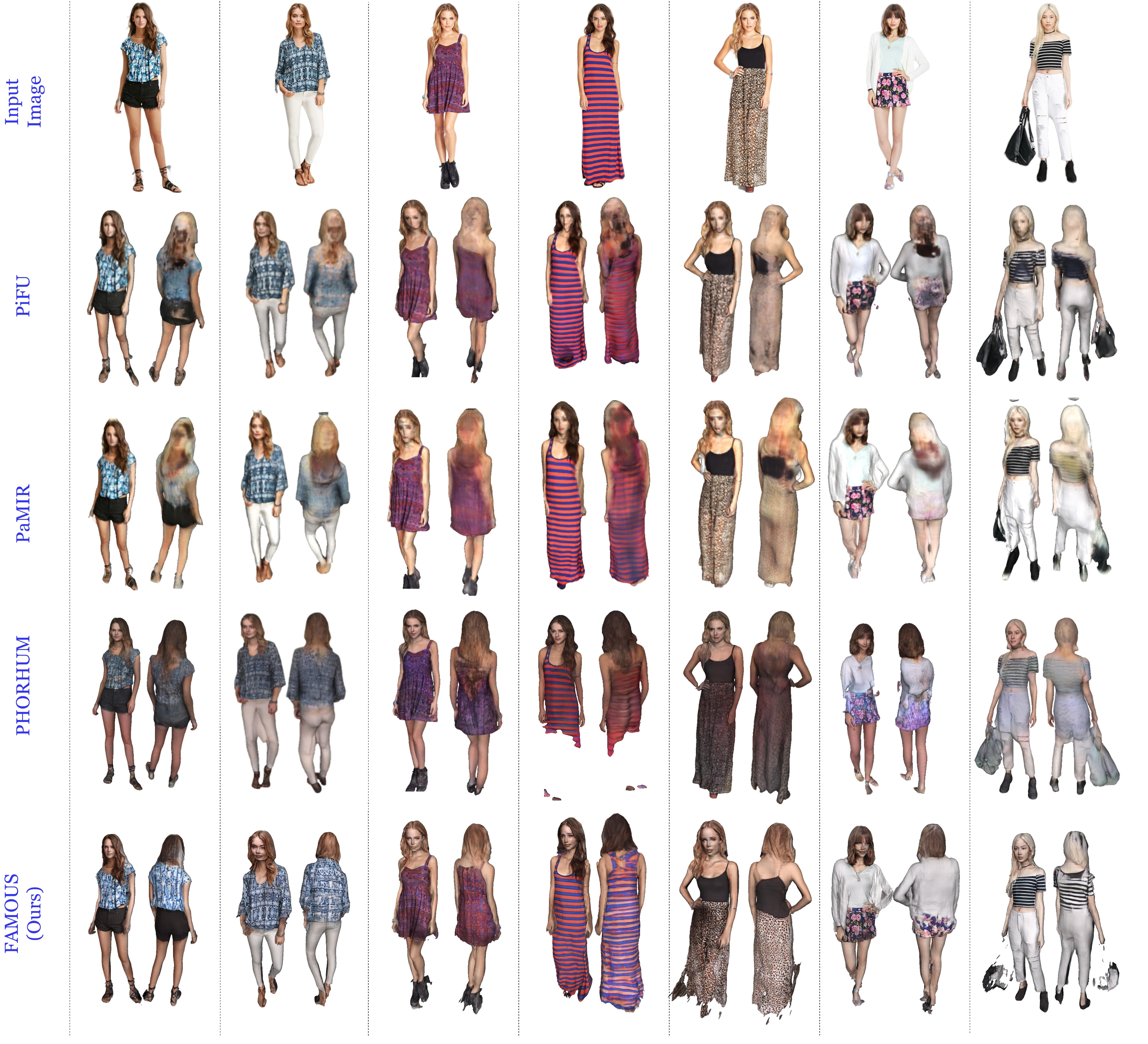}
\caption{Qualitative comparison of the SOTA pipelines predicting both shape and texture with ours
on the DeepFashion test set for the images shown on the top.
From top to bottom: Input image, PIFu \cite{pifu}, PaMIR \cite{pamir}, PHORHUM \cite{phorhum}, and FAMOUS (ours).
For each of these methods, we provide front-back slanted visualizations in order. FAMOUS visually outperforms these SOTA approaches in the occluded texture.}
\label{fig:supp-fig1}
\end{figure}

\section{Distributionally Aligned Hallucinator (DAH)}
\subsection{Implementation Details}

\begin{figure*}[!htbp]
\centering
\includegraphics[width=0.9\textwidth]{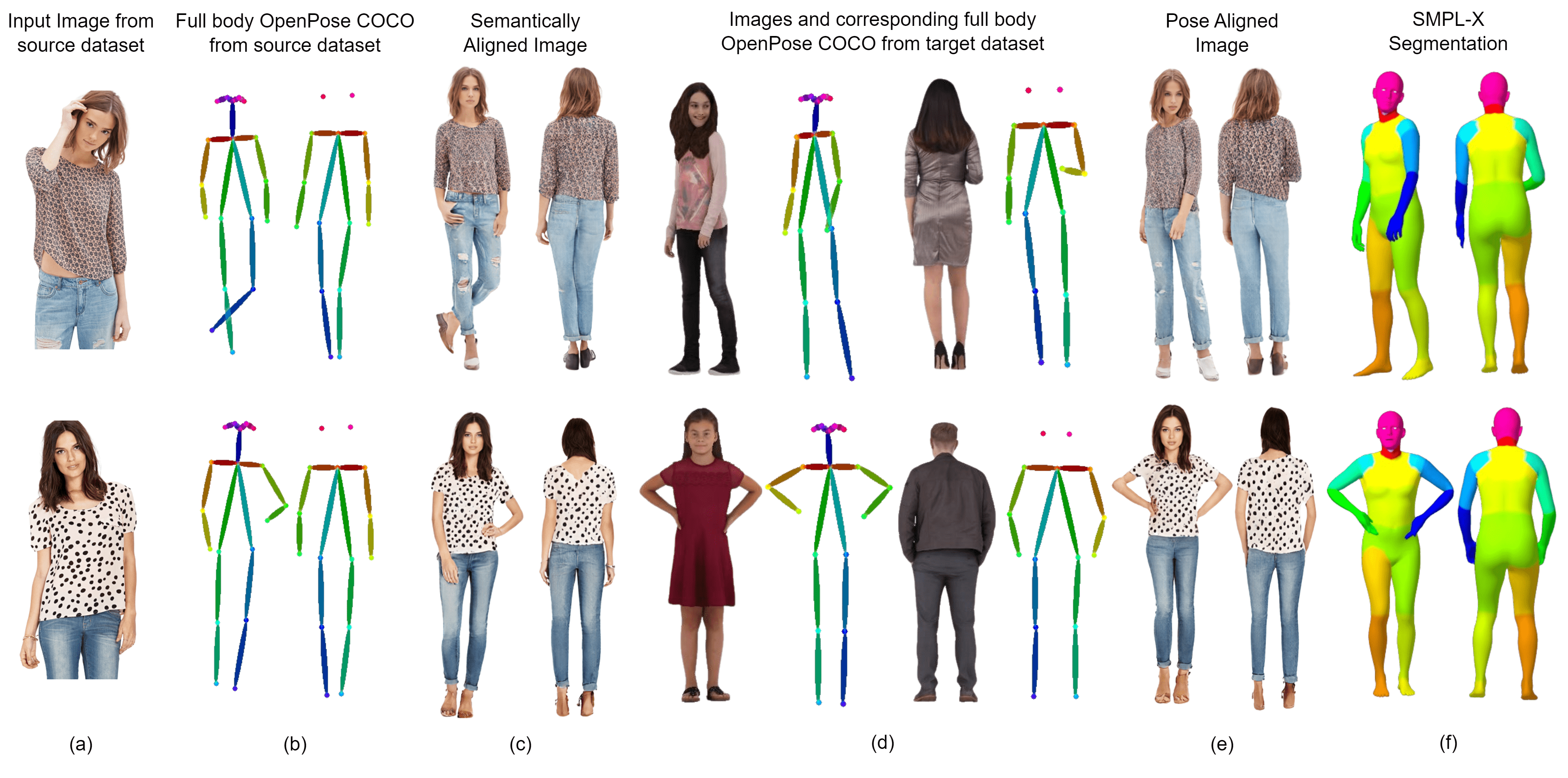}
\caption{A few data samples from the source DeepFashion dataset and corresponding data generated during each stage of alignment with the respective guidance used in the previous stage.
}
\label{fig:data_flow}
\end{figure*}

\noindent We follow the original training settings of the vanilla hallucinator \cite{nted} in this work.
However, this method was originally designed for the pose-guided view synthesis problem, where both the image and target OpenPose COCO skeleton are inputs. Therefore, at inference time, to obtain target SMPLX segmentation and the target silhouette, we had to fit SMPLX using \cite{pymafx} and segment the binary mask from the input image, respectively. The data flow used in each stage of alignment during the training phase is shown in Figure \ref{fig:data_flow}. For semantic alignment, we use the original deep fashion images as the image input. For guidance, we sample a subset of GT OpenPose COCO keypoints that represent full-body semantics provided in the original dataset. Using these sampled full body guidance, we generate front/back image pairs $\{ ( I^\mathcal{SA}_f, I^\mathcal{SA}_b ) \}$ as shown in Figure \ref{fig:data_flow}(c). These image pairs are then used to retrain the model in a self-supervised manner.

\begin{figure*}[!htbp]
\centering
\includegraphics[width=0.9\textwidth]{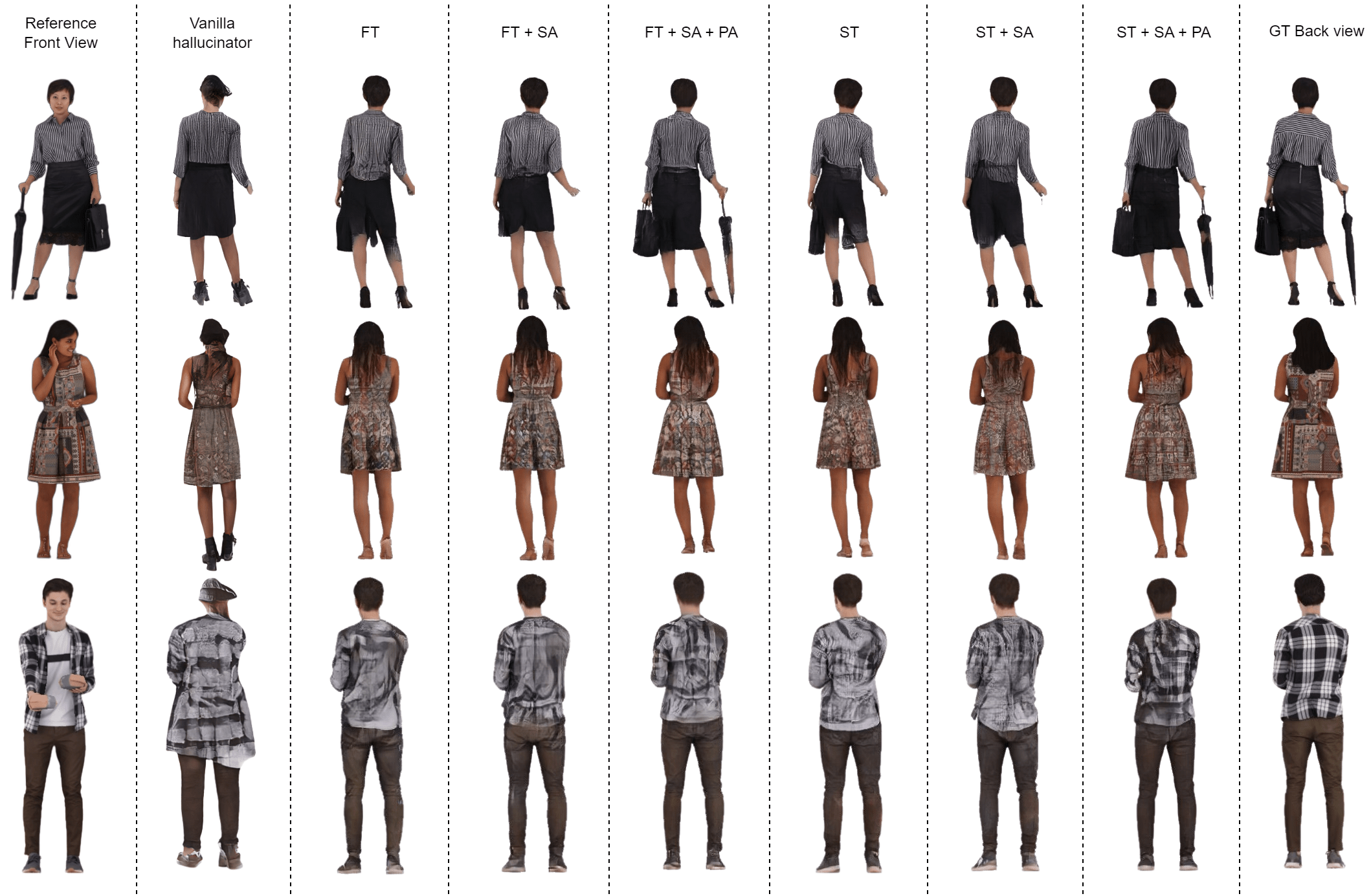}
\caption{Qualitative ablation of the disentangled distributional alignment for the 2D hallucinator. 
From left to right: Input image; Naive finetuning; Semantic Alignment with finetuning; Semantic and Pose Alignment with finetuning; Simultaneous training; semantic alignment with simultaneous training; semantic and pose alignment with simultaneous training, GT back view on target RenderPeople dataset.
}
\label{fig:alignement_ablation}
\end{figure*}

\par
\begin{wrapfigure}{11}{0.5\textwidth}
\centering
\vspace{-35pt}
\includegraphics[width=0.5\textwidth]{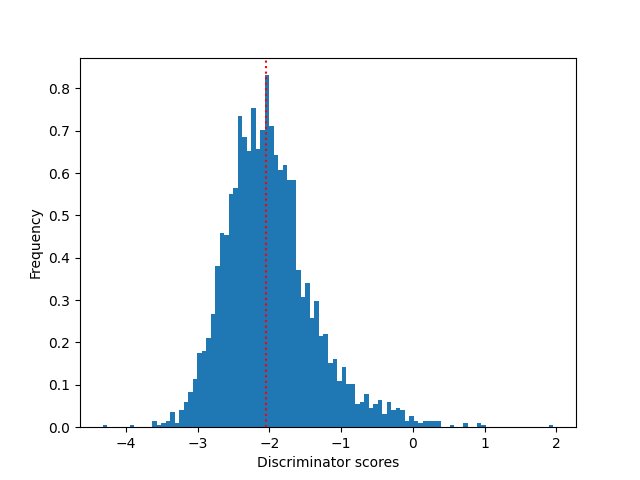}
\vspace{-10pt}
\caption{Histogram of discriminator scores obtained on the source deep fashion test set. The red dotted line denotes the median value of the distribution.
}
\label{fig:disc_score}
\end{wrapfigure}

 \noindent Following semantic alignment, we transfer the weights of the network to the pose alignment stage. During this stage, we use the collection of semantically aligned image pairs $\{ ( I^\mathcal{SA}_f, I^\mathcal{SA}_b ) \}$ as image input. For guidance, we sample a subset of GT OpenPose COCO keypoints from the target dataset, in our case, the RenderPeople dataset. The subset is collected based on a thresholding process using discriminator scores. For every guidance sample, we randomly select a few images from $\{ ( I^\mathcal{SA}_f, I^\mathcal{SA}_b ) \}$ and pass them through the network to obtain the corresponding outputs. Then we obtain the discriminator score (from vanilla discriminator checkpoints after vanilla hallucinator training \cite{nted}) of these outputs and threshold them to collect the subset. The threshold value is the median of discriminator scores obtained on the test set of the source deep fashion images. Figure \ref{fig:disc_score} shows the histogram of the discriminator scores and its median, denoted by a dotted red line. 

 \noindent Please note that the goal of this thresholding process is to maintain a high degree of realism and mitigate the artifacts generated in these image pairs $\{ ( I_f, I_b ) \}$. Using this subset of guidance, we generate front/back image pairs $\{ ( I^\mathcal{PA}_f, I^\mathcal{PA}_b ) \}$. Finally, we finetune the semantically aligned hallucinator with these pose-aligned pseudo pairs in a similar way described in the previous stage.

\par \noindent On a related note, we empirically find that off-the-shelf hallucinators often struggle to generate views for highly challenging poses. Therefore, it is advantageous to filter out these poses from our training set during pose alignment. If the target dataset includes poses that are more challenging than those in the 2D source dataset, we hypothesize that the information flow from our aligned 2D prior (in-domain to 3D) to the 3D target space will suffice for reliable textured reconstruction.

\par \noindent For style alignment, we checkpoint the network with weights after finetuning in the previous stage. At the beginning of this stage, we switch guidance to the SMPLX segmentation map and silhoutte. For the finetuning experiments, only the weights are transferred from the previous stages. But for simultaneous training, both weights and the pose-aligned generated image pairs are utilized. 

\vspace{-5pt}

\begin{table}[!htbp]
\begin{center}
\caption{Ablation study of the hallucinator on Render People.}
\label{tab:mmd}
\begin{adjustbox}{width=0.6\columnwidth,center}
\renewcommand{\arraystretch}{1.4}
\begin{threeparttable}
\begin{tabular}{l|cccc}
\hline
Experiments & FID($\downarrow$) & LPIPS($\downarrow$) & SSIM ($\uparrow$)& IS ($\uparrow$)\\ \hline
ST & 71.0900 & 0.0935 & 0.8678 & 2.7228\\
ST + Coral & 72.5376 & 0.0973 & 0.8652 & 2.5166\\
ST + MMD & 73.0580 & 0.0968 & 0.8651 & 2.5433\\
\hline
\end{tabular}
\begin{tablenotes}
\item [] \hspace{-0.4cm} ST $\equiv$ Simultaneous Training 
\end{tablenotes}
\end{threeparttable}
\renewcommand{\arraystretch}{1}
\end{adjustbox}
\end{center}
\end{table}

\vspace{-20pt}

\subsection{Experiment Details} 

\par \noindent \textbf{Effects of MMD and Coral}: Domain adaptation methods like Maximum Mean Discrepancy \cite{mmd} and Coral \cite{deep-coral} aim to reduce the distributional differences between the source and target datasets. But blindly minimizing this difference in the case of a generative model tends to confuse the network. Table \ref{tab:mmd} shows the quantitative ablation study for these experiments.

\par \noindent \textbf{Effects of each alignment}: An image can be represented based on semantics, pose, view, and style ~\cite{disentaglement}. For a conditional generative model like NTED \cite{nted}, we find semantics and pose distributional differences between two datasets play a crucial role in its alignment. Naively fine-tuning the vanilla hallucinator on the target dataset fails to generalize well due to these distributional differences between the datasets. Figure \ref{fig:alignement_ablation} shows the qualitative ablation for each stage of alignment.

\vspace{-0.15cm}
\section{Occupancy Network}
\subsection{Experiments} 

\begin{figure}[!htbp]
\centering
\includegraphics[width = 0.5\textwidth]{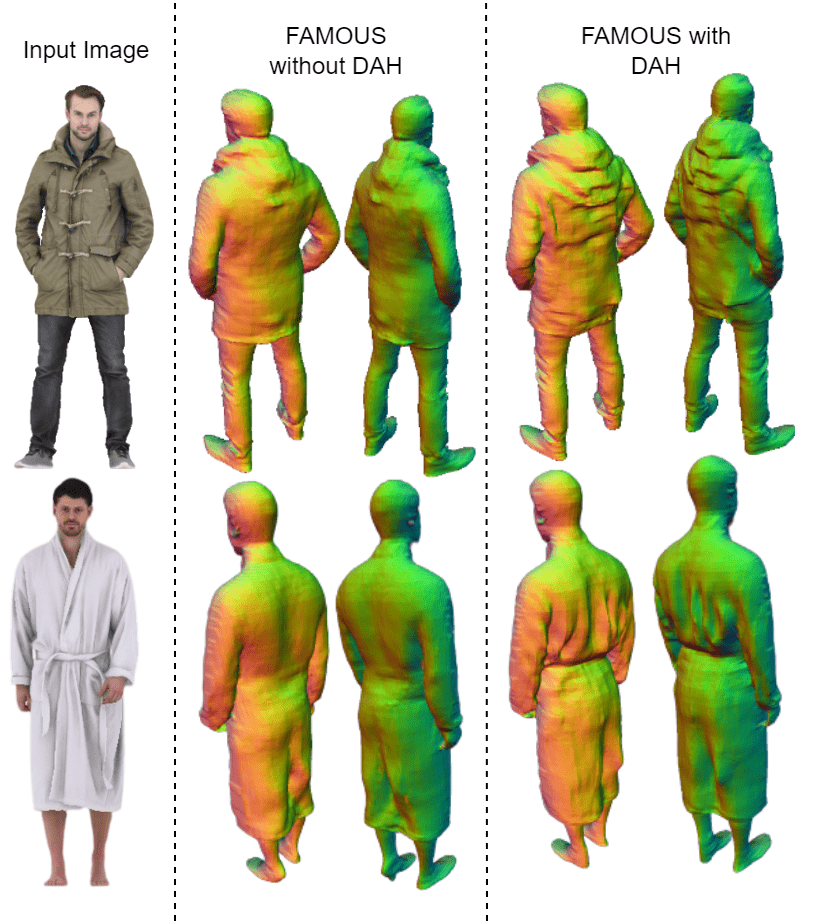}
\caption{\textbf{Qualitative ablation of shape prediction.} 
From left to right: Input image; The front/back surface normals from our framework without (middle)
and with (right) the aligned hallucinator prediction DAH. Note the high-precision details recovered in the estimated normals
after using aligned hallucinator prediction is compared to the vanilla case in the middle.
}
\label{fig:normal}
\end{figure}

\begin{figure*}[!htbp]
\centering
\includegraphics[width=\textwidth]{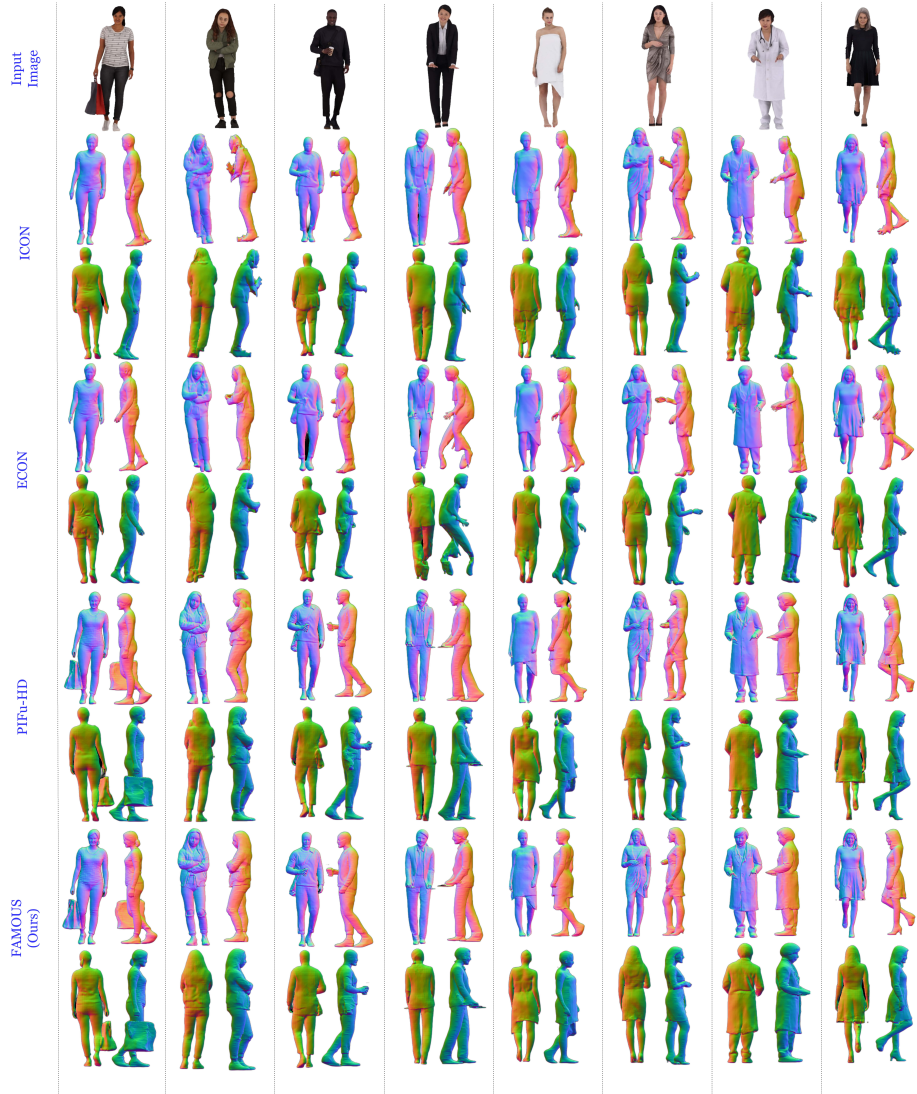}
\caption{
Qualitative comparison of ICON, ECON and PIFu-HD predicting shape with ours
on the RenderPeople test set for the images shown on the top.
From top to bottom: Input image, ICON \cite{icon}, ECON \cite{econ}, PIFu-HD \cite{pifuhd}, and FAMOUS (ours).
For each of these methods, we provide four visualizations: surface normals at yaw angle $0^{\circ}$, $90^{\circ}$, $180^{\circ}$ and $270^{\circ}$. FAMOUS visually outperforms these SOTA approaches in predicting surface details.}
\label{fig:econ_icon}
\end{figure*}

\noindent \textbf{Shape Analysis of ECON and ICON:} ICON and ECON take an RBG image and an estimated SMPL-X fit as input. However, recovering SMPL-X fit from a single image is still an unsolved problem, and errors in estimating SMPL-X fit are propagated to ECON and ICON. Furthermore, for fashion images, like in the renderpeople dataset, ICON tends to overfit to the SMPL-X mesh, as shown in \ref{fig:econ_icon}. ECON suffers from stitching artifacts and depth ambiguity compared to PIFu-HD \cite{econ}. So, we empirically find PIFu-HD to be the best occupancy network for fashion images and build our occupancy network on top of it. Figure \ref{fig:normal} shows the superior performance of the modified PIFu-HD implemented in our framework.

\vspace{-0.35cm}
\subsection{Evaluation metrics}
\vspace{-0.25cm}
\noindent We primarily evaluate FAMOUS with SOTA methods that predict surface colors: PIFu, PaMIR, and PHOROUM. We quantitatively evaluate texture quality by rendering the predicted textured mesh with the respective camera model by rotating the camera by [$0^\circ$, $90^\circ$, $180^\circ$, $270^\circ$] yaw angle and comparing it with corresponding GT renderings using image reconstruction metrics. 
For image reconstruction metrics, we report Learned Perceptual Image Patch Similarity (LPIPS), Inception Score (IS), structural similarity index (SSIM) and Frechet Inception Distance (FID).
For evaluating the reconstruction quality of geometry, we measure the difference between the reconstructed and GT meshes using Chamfer distance and Point to Surface (P2S). Additionally, we render the normal images between the predicted and GT mesh in the same way as colored images and report the MSE error.
Note that all the evaluations are performed on the original predicted meshes from the respective models. But for visualizations, the artifacts from predicted meshes are removed. Only the biggest connected, water-tight mesh is kept.

\begin{table}[!htbp]
\begin{center}
\caption{Ablation study of the Texture Completion on Render People.}
\label{tab:tex_pifu}
\begin{adjustbox}{width=0.65\columnwidth,center}
\renewcommand{\arraystretch}{1.4}
\begin{threeparttable}
\begin{tabular}{l|cccc}
\hline
Experiments & FID($\downarrow$) & LPIPS($\downarrow$) & SSIM ($\uparrow$)& IS ($\uparrow$)\\ \hline
Tex-PIFu* & 103.4353 & 0.1655 & 0.8953 & 3.4092\\
Tex-PIFu* + SIREN  & 78.3775 & 0.1593 & 0.9022 & 3.7334\\
FAMOUS& 55.0711 & 0.1425 & 0.9140 & 3.8655\\
\hline
\end{tabular}
\begin{tablenotes}
\item [] \hspace{-0.4cm} FAMOUS $\equiv$ Tex-PIFu* with SIREN and UNet backbone  
\item [] \hspace{-0.4cm} Tex-PIFu* $\equiv$ modified Tex-PIFu with predicted back view from DAH 
\end{tablenotes}
\end{threeparttable}
\renewcommand{\arraystretch}{1}
\end{adjustbox}
\end{center}
\end{table}

\section{Texture Completion}

\subsection{Implementation Details}

\par\noindent The backbone for the texture completion part is based on UNet Architecture 8 encoder-decoder layers with bilinear interpolation in between the filters. The first layer contains 64 channels, and then the channel size is doubled on each layer of the encoder. The decoder reduces the channel size by half on each layer. The final layer contains 128 channels. We use Leaky-ReLU activation for both the encoder and decoder. Once we extract features from both the front view and the generated back view, we project them onto the queried points in a pixel-aligned manner.
In comparison to Tex-PIFu, where the same pixel features from the 
encoders are used for color prediction of multiple 3D points along the line of projection (i.e. pixel-aligned). We only use pixel features of the corresponding view in each side. The MLP for predicting per-vertex color is similar to the implementation in Tex-PIFu except for activation at the last layer, which was changed to sine activation inspired by \cite{siren}. Table \ref{tab:tex_pifu} shows the quantitative ablation study for the texture completion part.

\begin{figure*}[!htbp]
\centering
  \includegraphics[height=0.65\linewidth]{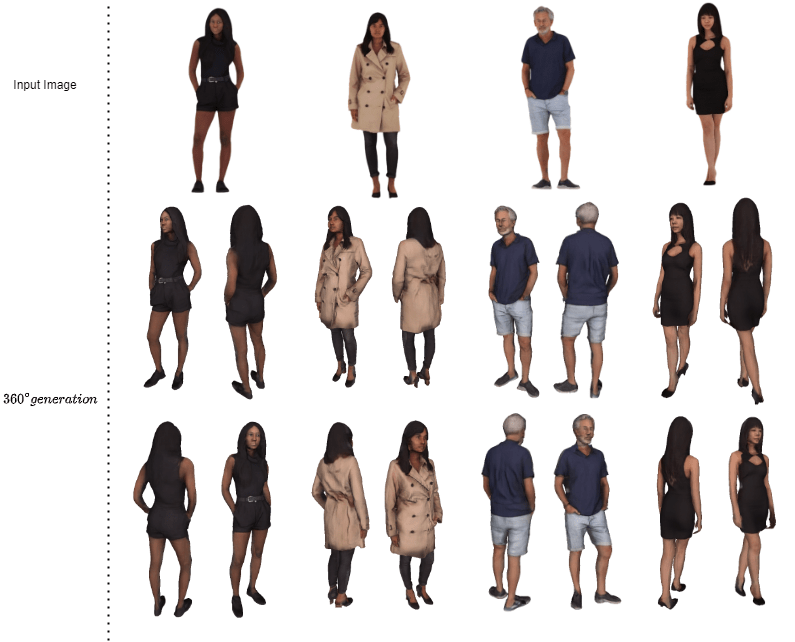}
   \caption{Additional $360^{\circ}$ visualization of our results on the RenderPeople test set.}
   \label{fig:sides}
\end{figure*}

\newpage
\bibliographystyle{splncs04}
\bibliography{
    bibliographies/misc,
    bibliographies/human3d, 
    bibliographies/gan2d,
    bibliographies/datasets,
    bibliographies/da
}